\documentclass[sigconf, nonacm, pdfa]{acmart}

\usepackage[a-2b]{pdfx}

\usepackage{newtxtext}
\usepackage{upgreek}
\usepackage{bbding}
\usepackage{makecell}
\usepackage{booktabs}
\usepackage{ragged2e} 
\usepackage{multirow}
\usepackage{balance}

\usepackage{algorithmic}
\usepackage{algorithm}
\usepackage{enumitem}

\newcommand{\stitle}[1]{\noindent{\bf #1\/}}

\newcommand\vldbdoi{10.14778/3705829.3705830}
\newcommand\vldbpages{80 - 92}
\newcommand\vldbvolume{18}
\newcommand\vldbissue{2}
\newcommand\vldbyear{2024}
\newcommand\vldbauthors{\authors}
\newcommand\vldbtitle{\shorttitle} 
\newcommand\vldbavailabilityurl{https://github.com/slzhou-xy/RED}
\newcommand\vldbpagestyle{empty} 

\begin{document}
\title{RED: Effective Trajectory Representation Learning with Comprehensive Information}

\author{Silin Zhou}
\affiliation{
  \institution{University of Electronic Science and Technology of China}
}
\email{zhousilinxy@gmail.com}

\author{Shuo Shang}
\authornote{The corresponding author.}
\affiliation{
  \institution{University of Electronic Science and Technology of China}
}
\email{jedi.shang@gmail.com}

\author{Lisi Chen}
\affiliation{
  \institution{University of Electronic Science and Technology of China}
}
\email{lchen012@e.ntu.edu.sg}

\author{Christian S. Jensen}
\affiliation{
 \institution{Aalborg University}
}
\email{csj@cs.aau.dk}

\author{Panos Kalnis}
\affiliation{
  \institution{KAUST}
}
\email{panos.kalnis@kaust.edu.sa}



\begin{abstract}
Trajectory representation learning (TRL) maps trajectories to vectors that can then be used for various downstream tasks, including trajectory similarity computation, trajectory classification, and travel-time estimation. However, existing TRL methods often produce vectors that, when used in downstream tasks, yield insufficiently accurate results. A key reason is that they fail to utilize the comprehensive information encompassed by trajectories. We propose a self-supervised TRL framework, called RED, which effectively exploits multiple types of trajectory information. 
Overall, RED adopts the Transformer as the backbone model and masks the constituting paths in trajectories to train a masked autoencoder (MAE). 
In particular, RED considers the moving patterns of trajectories by employing a \textit{\textbf{R}oad-aware masking strategy} that retains key paths of trajectories during masking, thereby preserving crucial information of the trajectories. RED also adopts a \textit{spatial-temporal-user joint \textbf{E}mbedding} scheme to encode comprehensive information when preparing the trajectories as model inputs. To conduct training, RED adopts \textit{\textbf{D}ual-objective task learning}: the Transformer encoder predicts the next segment in a trajectory, and the Transformer 
decoder reconstructs the entire trajectory. RED also considers the spatial-temporal correlations of trajectories by modifying the attention mechanism of the Transformer. 
We compare RED with 9 state-of-the-art TRL methods for 4 downstream tasks on 3 real-world datasets, finding that RED can usually improve the accuracy of the best-performing baseline by over 5\%.
\end{abstract}

\maketitle

\pagestyle{\vldbpagestyle}
\begingroup\small\noindent\raggedright\textbf{PVLDB Reference Format:}\\
\vldbauthors. \vldbtitle. PVLDB, \vldbvolume(\vldbissue): \vldbpages, \vldbyear.\\
\href{https://doi.org/\vldbdoi}{doi:\vldbdoi}
\endgroup
\begingroup
\renewcommand\thefootnote{}\footnote{\noindent
This work is licensed under the Creative Commons BY-NC-ND 4.0 International License. Visit \url{https://creativecommons.org/licenses/by-nc-nd/4.0/} to view a copy of this license. For any use beyond those covered by this license, obtain permission by emailing \href{mailto:info@vldb.org}{info@vldb.org}. Copyright is held by the owner/author(s). Publication rights licensed to the VLDB Endowment. \\
\raggedright Proceedings of the VLDB Endowment, Vol. \vldbvolume, No. \vldbissue\ %
ISSN 2150-8097. \\
\href{https://doi.org/\vldbdoi}{doi:\vldbdoi} \\
}\addtocounter{footnote}{-1}\endgroup

\ifdefempty{\vldbavailabilityurl}{}{
\vspace{.3cm}
\begingroup\small\noindent\raggedright\textbf{PVLDB Artifact Availability:}\\
The source code, data, and/or other artifacts have been made available at \url{https://github.com/slzhou-xy/RED}.
\endgroup
}

\section{Introduction} 
\label{Intro}

With the proliferation of GPS-enabled devices (e.g., smartphones, navigators, and digital watches), large amounts of trajectories are collected, recording the movements of pedestrians or vehicles. These trajectories serve as the foundation for many applications such as traffic prediction~\cite{zhao2023traffic_prediction}, urban planning~\cite{pedersen2020urban_planning}, and transportation optimization~\cite{yang2021transportation_optimization}. However, as sequences of timestamped locations, trajectories require specific techniques for management and analysis. For example, trajectory similarity computation often relies on dynamic programming~\cite{dtw,edr}, resulting in computational costs that increase quadratically with trajectory length.

Recently, \textit{trajectory representation learning} (TRL), which maps each trajectory to a vectors embedding, has attracted attention as a general preprocessing technique~\cite{mao2022JCLRNT,yang2022wsccl,jiang2023start,lightpath}. The advantage is that the learned vectors can be used 
directly in many downstream tasks, including trajectory classification~\cite{Trajformer}, travel time estimation~\cite{tte1}, and trajectory similarity computation~\cite{grlstm}, with standard vector processing techniques. For instance, trajectory similarity can be calculated as the distance between the vectors of two trajectories, with a cost that does not increase with trajectory length.

Early TRL methods~\cite{traj2vec,t2vec} usually target a specific task. For example, Traj2vec~\cite{traj2vec} uses Recurrent Neural Networks (RNNs) to map trajectories to vectors and tailors model design for trajectory clustering task. The problem of these methods is that their vectors do not work well when used in other downstream tasks. Subsequent TRL methods~\cite{trembr,pim,toast} utilize self-supervised learning (SSL) due to the strong generalization capabilities of SSL. In particular, SSL relies on a generic pre-training task to distill data information, e.g., by manually masking a portion of the trajectories and learning to recover the trajectories. For example, Trembr~\cite{trembr} uses RNNs~\cite{lstm} with an encoder-decoder architecture. The encoder embeds each trajectory into a vector, and the decoder recovers the trajectory from this vector. Recent TRL methods~\cite{mao2022JCLRNT,jiang2023start,lightpath} adopt contrastive learning (CL). CL conducts data augmentations to generate positive and negative samples for each trajectory, and the model is trained to make positive trajectory pairs more similar than negative pairs. For instance, START~\cite{jiang2023start} features several data augmentation techniques, such as randomly masking segments in the trajectories and trimming the trajectories to get sub-trajectories. However, data augmentation techniques may not generalize across datasets because their performance depends strongly on the dataset under consideration. 

\begin{table}[!htbp]
    \LARGE
   \centering
   \caption{Trajectory information utilized by TRL methods.}
   \label{tab:trl_arch}
   \resizebox{\linewidth}{!}{
   \begin{tabular}{l|c|c|c|c|c|c}
   \toprule
      \textbf{Model} & \textbf{Road} & \textbf{User} & \textbf{Spatial} & \textbf{Temporal}  & \textbf{Travel} & \textbf{Movement} \\ 
      \midrule
      T2vec~\cite{t2vec} & \XSolidBold & \XSolidBold &\XSolidBold & \XSolidBold  & \XSolidBold & \XSolidBold \\
      Traj2vec~\cite{traj2vec} & \XSolidBold & \XSolidBold &\XSolidBold & \XSolidBold  & \XSolidBold & \CheckmarkBold \\
      Trembr~\cite{trembr} & \CheckmarkBold & \XSolidBold &\XSolidBold & \CheckmarkBold  & \XSolidBold & \XSolidBold  \\
      PIM~\cite{pim} & \CheckmarkBold & \XSolidBold &\CheckmarkBold & \XSolidBold  & \XSolidBold & \XSolidBold \\
      Toast~\cite{toast} & \CheckmarkBold & \XSolidBold &\CheckmarkBold & \XSolidBold  & \CheckmarkBold & \CheckmarkBold  \\
      JCLRNT~\cite{mao2022JCLRNT} & \CheckmarkBold & \XSolidBold &\CheckmarkBold & \XSolidBold  & \XSolidBold & \XSolidBold \\
      START~\cite{jiang2023start} & \CheckmarkBold & \CheckmarkBold &\CheckmarkBold & \CheckmarkBold  & \CheckmarkBold  & \XSolidBold \\
      LightPath~\cite{lightpath} & \CheckmarkBold & \XSolidBold &\XSolidBold & \XSolidBold  & \CheckmarkBold & \XSolidBold  \\
      RED (\textbf{ours}) & \CheckmarkBold &\CheckmarkBold & \CheckmarkBold & \CheckmarkBold & \CheckmarkBold & \CheckmarkBold \\
      \bottomrule
   \end{tabular}
   }
\end{table}

A trajectory encompasses multiple kinds of information, including road, user, spatial-temporal, travel, and movement, but existing TRL methods do not utilize them comprehensively, resulting in trajectory representation vectors with reduced accuracy. In particular, \textit{road information}, obtained by map matching~\cite{fmm}, tells the road segments each trajectory passes and is less noisy than the raw trajectory points.  
\textit{User information} ties each trajectory to a user and encodes user preference. \textit{Spatial information} refers to the spatial locations of the road segments, and a road segment usually has multiple adjacent road segments with high transition probabilities~\cite{gts}. \textit{Temporal information} captures the progress over time when moving along a trajectory and can be used to estimate travel time and analyze traffic flow~\cite{jiang2023start}. \textit{Travel semantics}~\cite{toast} refers to the overall travel statistics of a trajectory, such as travel time and distance. \textit{Movement semantics}~\cite{traj2vec} refers to the local properties of trajectories, such as whether congestion is encountered at a location.

Table~\ref{tab:trl_arch} summarizes how trajectory information is utilized by existing TRL methods. Road information is ignored by a few methods, e.g., Traj2vec~\cite{traj2vec} models trajectories as point sequences without considering the underlying road properties. Spatial information is usually considered by using the road network. For example, Toast~\cite{toast} and PIM~\cite{pim} apply the graph neural network to capture the topology of the road network of a city. Temporal information is typically used to encode model inputs, e.g., Trembr~\cite{trembr} includes the timestamp of each road in a trajectory as input to its RNN. Some methods incorporate the travel semantics into trajectories to extract global information. For instance,  Toast~\cite{toast}, START~\cite{jiang2023start}, and LightPath~\cite{lightpath} utilize the learned trajectory representations to recover raw trajectories. Only a few methods consider the moving semantics for local information, e.g., Traj2vec and Toast use context windows to extract short sequences from trajectories. START~\cite{jiang2023start} is the only existing method that exploits user information.

To go beyond existing studies, we propose RED, a self-supervised learning method that aims to improve TRL by using more comprehensive trajectory information. For overall architecture, RED adopts Transformer as the backbone model and masks the segments of path trajectories to train a masked autoencoder. Regarding trajectory information, we first consider and analyze the \textit{local driving pattern} of \textit{road} path trajectories and propose a novel road-aware masking strategy. In particular, we split a path trajectory into the key path and the mask path, where the key path contains most of the semantic information of a trajectory to assist in the final trajectory representation. Then, we design a spatial-temporal-user joint embedding scheme to fuse \textit{spatial}, \textit{temporal}, and \textit{user} information of road paths for model inputs. We also co-encode the \textit{time} and \textit{segment type} to capture periodic information of travel. To train RED, we utilize two training tasks, including next segment prediction and trajectory reconstruction. Next segment prediction predicts the next road segment from previous key paths of a trajectory, while trajectory reconstruction reconstructs the original trajectory using the learned representations of the key paths and mask paths. Both tasks exploit the \textit{travel} information in the trajectory representations. Moreover, we also introduce two strategies to improve trajectory modeling: virtual tokens and a spatial-temporal correlation enhanced attention module. Virtual tokens solve the path misalignment issues when using the Transformer for trajectories, and the attention module enhances Transformer attention to extract the spatial-temporal information of trajectories.

To evaluate RED, we conduct experiments on 3 real-world datasets and compare with 9 state-of-the-art TRL methods, includes supervised and self-supervised methods, as well as 7 non-learning trajectory similarity computation methods. 
We consider 4 popular downstream tasks: travel time estimation, trajectory classification, trajectory similarity computation, and most similar trajectory retrieval. The results show that RED outperforms all existing methods in terms of accuracy across the tasks and datasets. In particular, compared with the best-performing baseline, the average accuracy improvement of RED at travel time estimation, trajectory classification, and trajectory similarity computation are 
7.03\%, 12.11\%, and 20.02\%, respectively. An ablation study suggests that all the types of trajectory information we utilize contribute to the accuracy of TRL and that our model designs are effective.

To summarize, we make three main contributions.
\begin{itemize}[leftmargin=*]
    \item We identify design limitations of existing TRL methods, including the use of random masking, insufficient utilization of trajectory information, and limited supervision signals for model training.  
    \item We propose RED as a more effective self-supervised learning framework for TRL. RED features a road-aware mask, spatial-temporal-user embedding, and dual-objective learning to address the limitations of existing methods.  
    \item We report on an extensive evaluation and comparison of RED with state-of-the-art TRL methods. The results show that RED is capable of higher accuracy than the existing methods across important downstream tasks. 
\end{itemize}

\section{Problem and Background} \label{preliminaries}
\subsection{Preliminaries}\label{def:rp} 
\noindent
\textbf{Definition 1 (GPS Trajectory).} A GPS trajectory, denoted as $\mathcal{T}_{gps}$, is a sequence of points collected at a fixed sampling time interval rate. Each point in $\mathcal{T}_{gps}$ takes the form of  $p_i = (x_i, y_i, t_i)$, where $x_i$, $y_i$, and $t_i$ denote longitude, latitude, and timestamp, respectively.

\begin{figure*}[!t]
    \centering
    \includegraphics[interpolate=False,width=0.95\linewidth]{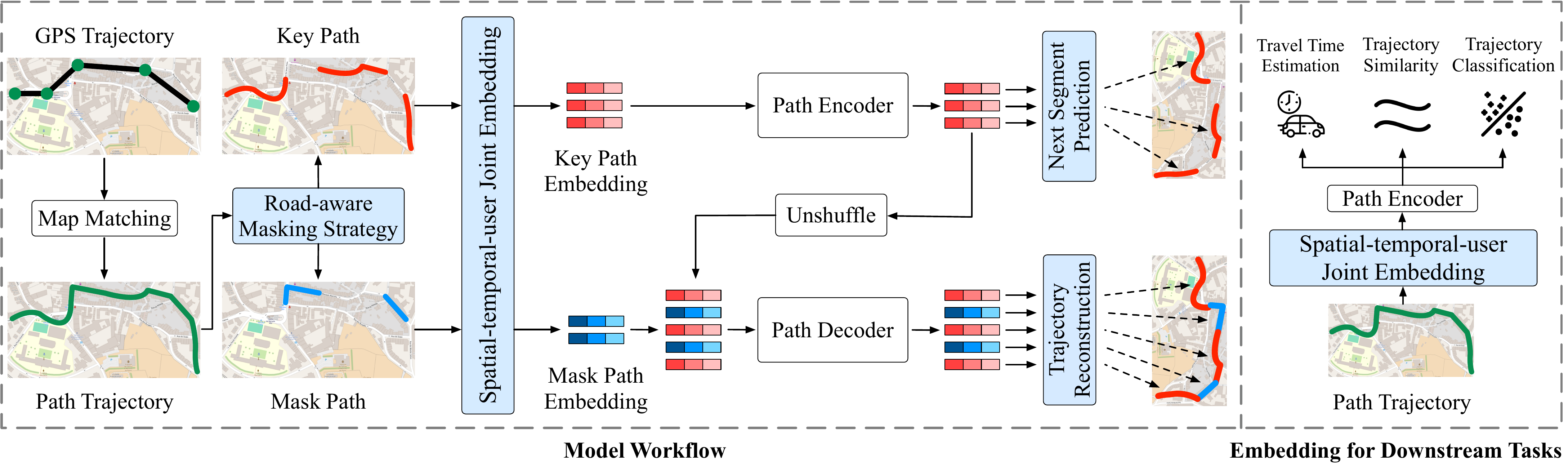}
    \caption{Overall architecture of the RED framework.}
    \label{fig:overview}
\end{figure*} 

\stitle{Definition 2 (Road Network).} A road network is modeled as a directed graph $\mathcal{G} = ( \mathcal{V}, \mathcal{A})$, where $ \mathcal{V} $ denotes the road segment set in the road network, and $\mathcal{A} \in \mathbb{R}^{\vert \mathcal{V} \vert \times \vert \mathcal{V} \vert}$ is the adjacency matrix that represents the connectivity between road segments. $\mathcal{A}[i, j] = 1$ if and only if road segments $v_i$ and $v_j$ are directly connected, otherwise $\mathcal{A}[i, j] = 0$. Under this definition, a trajectory can be extracted as a sequence of road segments that it passes through.

\stitle{Definition 3 (Path Trajectory).} A path trajectory $\mathcal{T}$ is a time-ordered sequence of road segments that is generated from $\mathcal{T}_{gps}$ by map matching. That is, $\mathcal{T}=\left< \tau_1, \tau_2, ..., \tau_{|\mathcal{T}|} \right>$ contains the $|\mathcal{T}|$  road segments passed by $\mathcal{T}_{gps}$. Each element $ \tau_i = (v_i, t_i) \in \mathcal{T}$ models that the trajectory passes road segment $v_i$ at timestamp $t_i$.

\subsection{Problem Statement}
Given a set of path trajectories $\mathcal{D} = \left\{\mathcal{T}_1, \mathcal{T}_2, ..., \mathcal{T}_{|\mathcal{D}|} \right\}$ and a road network graph $\mathcal{G}$, we aim to compute a generic vector representation $\vec{p}_i \in \mathbb{R}^{l}$ for each path trajectory $\mathcal{T}_i \in \mathcal{D}$, where $l$ is the dimension of the trajectory vector representation. We expect these vectors to achieve a high accuracy for various downstream tasks:
\begin{itemize}[leftmargin=*]
    \item \textbf{Trajectory similarity computation}: Given trajectories $\mathcal{T}_a$ and $\mathcal{T}_b$, trajectory similarity computation calculates a score capturing the similarity between $\mathcal{T}_a$ and $\mathcal{T}_b$.
    \item \textbf{Most similar trajectory retrieval}: Given a query trajectory $\mathcal{T}_a$ and a trajectory dataset $\mathcal{D}$, this task finds the trajectory $\mathcal{T}_b \in \mathcal{D}$ that is the most similar to $\mathcal{T}_a$.
    \item \textbf{Trajectory classification}: Given a trajectory $\mathcal{T}_a$, this task assigns $\mathcal{T}_a$ to a category, e.g., a user ID. 
    \item \textbf{Travel time estimation}: Given trajectory $\mathcal{T}_a$ without temporal information, this task predicts the travel time of $\mathcal{T}_a$. 
\end{itemize}

\subsection{Background on Machine Learning}
\noindent
\textbf{Transformer.} The Transformer~\cite{transformer} architecture has demonstrated impressive effectiveness at NLP. It models a token sequence, where a \textit{token} is a word identifier, that can be converted into a learnable vector. A token does not need to be associated with a word. Thus, a language model BERT~\cite{bert} designs a [CLS] token to represent the summary of a sentence.
The Transformer architecture encompasses multiple stacked blocks, each block contains a multi-head self-attention and a feed-forward network. Multi-head self-attention learns the inter-relationships between different elements in a sequence. A feed-forward network further enhances feature extraction and the model's expressiveness. Compared to RNN-based methods~\cite{lstm}, the Transformer can accommodate billions of model parameters by stacking blocks and computes a long token sequence in parallel, without the need to iterate through time steps.

An input token sequence embedding is given by $\mathbf{X} \in \mathbb{R}^{n \times l}$, where $n$ is the sequence length, and $l$ is the dimensionality of vectors. The vanilla Transformer first adds a position encoding to $\mathbf{X}$ and then transforms $\mathbf{X}$ to a query matrix $Q = \mathbf{X} W^Q \in \mathbb{R}^{n \times l}$, a key matrix $K = \mathbf{X} W^K \in \mathbb{R}^{n \times l}$, and a value matrix $V = \mathbf{X} W^V \in \mathbb{R}^{n \times l}$, $W^Q \in \mathbb{R}^{l \times l}$, $W^K \in \mathbb{R}^{l \times l}$, and $W^V \in \mathbb{R}^{l \times l}$ 
are learnable matrices. Next a self-attention correlation matrix is given by $\mathbf{A} = \frac{Q K^T}{\sqrt{l}} \in \mathbb{R}^{n  \times n}$, where $\alpha_{i,j} \in \mathbf{A}$ is the attention score between element-$i$ and element-$j$ of $\mathbf{X}$. The sequence output is computed as follows:
\begin{equation}
    \mathbf{X}^\prime = \textrm{softmax} ( \mathbf{A}) V,
    \label{eq:vanilla_sa}
\end{equation}
where $\mathbf{X}^\prime$ is the input to the subsequent feed-forward network. 

\stitle{Masked Autoencoder.} A masked autoencoder~\cite{mae} is an encoder-decoder model from computer vision (CV) that uses the Transformer. An image is first divided into small patches, where each patch can be seen as a word and can be placed in the Transformer for training. A masked autoencoder randomly masks partial patches and has its encoder act on visible patches. A lightweight decoder is used to reconstruct the raw image pixels based on mask tokens and potential representations from the encoder. This reconstruction task is an effective and meaningful self-supervised task.

In our study, we view a path trajectory as a sentence, and we view each segment as a word that is transformed into a token. Our base model is a masked autoencoder. We use prior trajectory knowledge to partition mask data, and we introduce comprehensive trajectory information to improve the accuracy of trajectory representation.
\section{Method Overview}\label{overview}

The left plot of Figure~\ref{fig:overview} shows the workflow of RED during training, which involves three key modules, i.e., \textit{road-aware masking}, \textit{spatial-temporal-user joint embedding}, and \textit{dual-objective learning}.

\stitle{Road-aware Masking.} As the first step, the map matching algorithm~\cite{fmm} is conducted to transform a GPS trajectory to a path trajectory. Then, the road-aware masking strategy splits a path trajectory $\mathcal{T}$ into \textit{key path set} $\mathcal{T}^k$ and \textit{mask path set} $\mathcal{T}^m$, where $\mathcal{T} = \mathcal{T}^k \cup \mathcal{T}^m$, based on the sampling rate and driving pattern of the trajectory. The key paths in $\mathcal{T}^k$ encompass the driving patterns of the trajectory while the mask paths in $\mathcal{T}^m$ are less crucial and thus are masked for the encoder. In comparison, existing methods adopt the random masking strategy, which randomly masks the paths. Trajectory information will be lost if some key paths are masked. Moreover, random masking also requires extensive tuning of the mask ratio while our road-aware masking does not.

\stitle{Joint Embedding.} Next, each segment of a trajectory is encoded as a raw embedding to serve as framework input, which can be expressed as $\mathbf{X} = \textrm{Emb}(\mathcal{T})$. The spatial-temporal-user joint embedding integrates comprehensive information of the trajectory, including spatial features from the road network graph, temporal features from the traffic and travel patterns over time, user features from user ID, and segment features from the road types. 

\stitle{Dual-objective learning.} We train RED with two objectives on to provide sufficient supervision signals for learning. In particular, the encoder maps the key paths in $\mathcal{T}^k$ to key path embedding $\widehat{\mathbf{X}}^k = \textrm{Encoder}(\mathbf{X}^k)$.  As the key paths capture the crucial driving patterns of a trajectory, we use the next segment prediction objective $\mathcal{L}^{nsp}$ for the encoder to predict the next key path given the previous key paths. This resembles next token prediction in NLP. The decoder takes the path embeddings generated by the encoder, merges the key paths and mask paths according to their original order in the trajectory, and reconstructs the trajectory using the embeddings, i.e., $\widehat{\mathbf{X}}^\mathcal{T} = \textrm{Decoder}(\textrm{Unshuffle}(\widehat{\mathbf{X}}^k, \mathbf{X}^m))$. Thus, we use the trajectory reconstruction objective $\mathcal{L}^{tr}$ to train the decoder. Therefore, the overall objective function is:
\begin{equation}
  \mathcal{L} = \lambda_1 \mathcal{L}^{nsp}(\widehat{\mathbf{X}}^k) + (1 - \lambda_1) \mathcal{L}^{tr}(\widehat{\mathbf{X}}^\mathcal{T}) 
  \label{eq:loss},
\end{equation}
where $\lambda_1$ controls the weight of the loss terms. To tackle sparsity on trajectories, i.e., GPS points recorded by the device during the driving process are discontinuous, inconsistent, or missing, we incorporate comprehensive information, more effective supervision signals, and a GNN for the road network.

As shown to the right in Figure~\ref{fig:overview}, during inference, we only utilize the path encoder and feed the embeddings of the complete trajectory as input to get the trajectory vector for downstream tasks.

\begin{figure}[!t]
    \centering
    \includegraphics[width=\linewidth,interpolate=False]{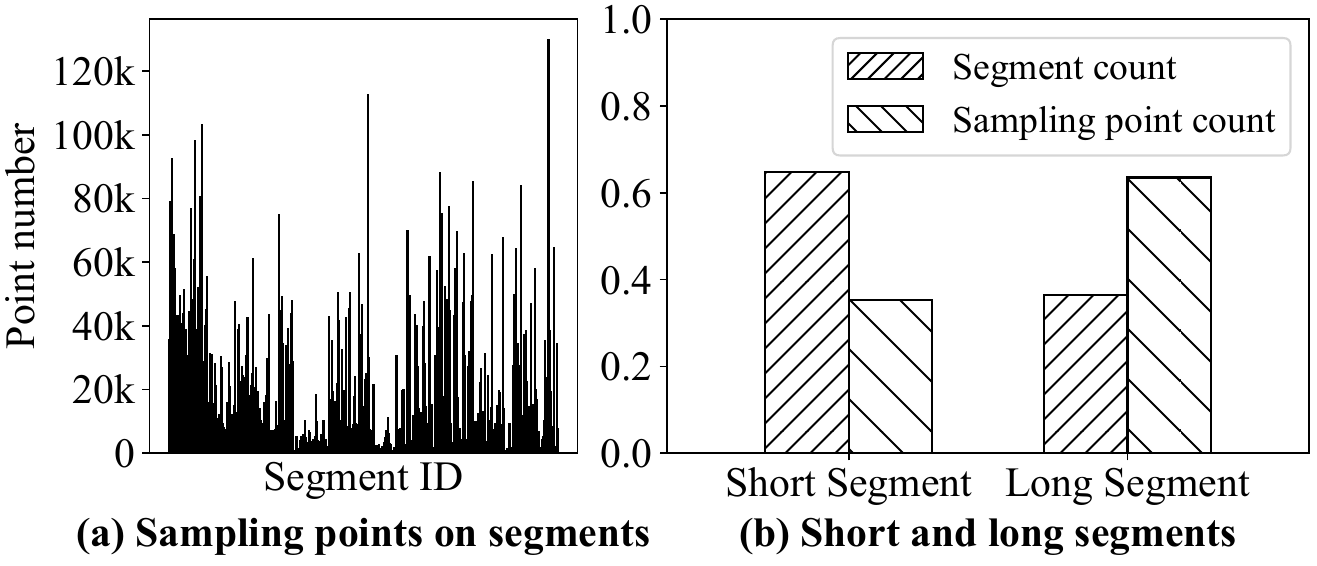}
    \caption{Trajectory sample statistics of the Porto dataset.}
    \label{fig:statistics1}
\end{figure}

\section{Key Designs of RED} \label{method}
We proceed to detail the key innovations of RED, which includes the road-aware masking strategy, the spatial-temporal-user joint embedding, the dual-objective task learning, and the enhanced trajectory modeling techniques.

\subsection{Road-aware Masking Strategy} \label{sec:road-aware mask}
Mask-based self-supervised learning is popular for TRL, which first masks some road segments of a path trajectory, and then learns to reconstruct the masked road segments. However, existing methods use the random masking strategy, which has two problems. 
First, it may discard essential road segments of a trajectory, leading to some key information loss and inaccurate trajectory representations.
Second, the masking ratio requires extensive tuning, and one masking ratio may not suit different trajectories. We introduce a \textit{road-aware masking strategy} that avoids the two problems. 

To motivate the road-aware masking strategy, we illustrate the sampling statistics of real-world trajectories in Figure~\ref{fig:statistics1}. The dataset is Porto, and the sampling interval is about 15 seconds for consecutive points in a trajectory. Since the average length of the road segments in the road network is about 85 meters, we call road segments below 85 meters as short segments and the converse long segments. Figure~\ref{fig:statistics1}(a) shows that the distribution of the sampling points over the road segments is highly skewed, i.e., some segments have many sampling points while some segments have only a few sampling points. Intuitively, segments with many sampling points are important because they are passed by many trajectories and may take an important role in the road network (e.g., connecting two important areas). Figure~\ref{fig:statistics1}(b) shows that there are more short road segments than long road segments in the road network but more sampling points reside on the long road segments than the short segments. This suggests that the long segments are more important than the short segments because the trajectories spend more time on them. Overall, Figure~\ref{fig:statistics1} suggests that the road segments are different in their importance. Thus, by treating all road segments equally, random masking may lose important segments and hinder model training.   

\begin{figure}[!t]
    \centering
    \includegraphics[width=0.9\linewidth,interpolate=False]{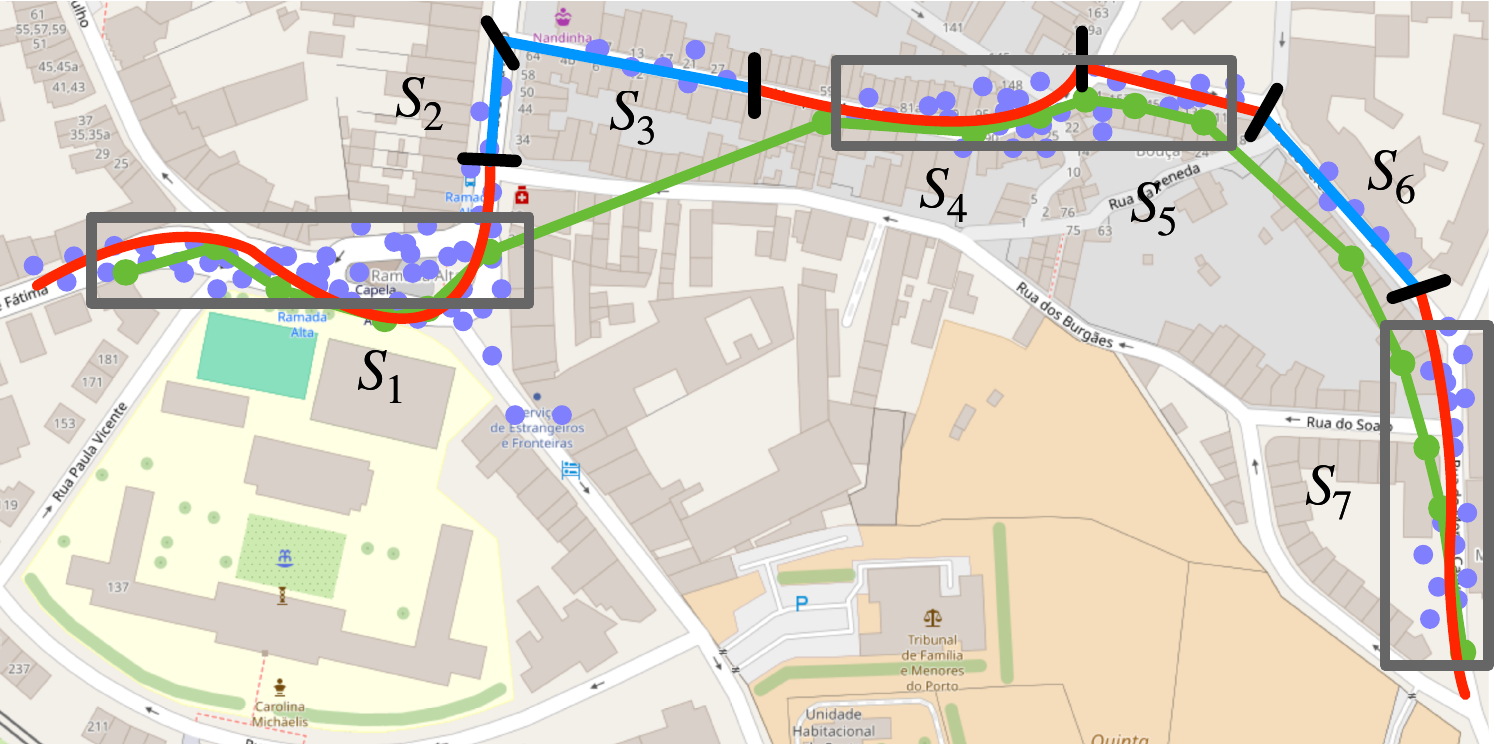}
    \caption{Illustration of the road-aware masking.}
    \label{fig:adaptive}
\end{figure}

Driven by the observations from Figure~\ref{fig:statistics1}, our road-aware masking strategy treats road segments with many sampling points and long road segments as \textit{key paths}, which will not be masked. To avoid parameter tuning, we call a segment hot (resp. long) when its sampling points (resp. length) are larger than the average over the road segments. Figure~\ref{fig:adaptive} provides a running example of road-aware masking. The original trajectory is plotted in green color. After map matching, we obtain 7 road segments, i.e., $\left< s_1, s_2, ..., s_7 \right>$. The key paths, i.e., $(s_1$, $s_4$, $s_5$, $s_7)$, are marked in red, while the mask paths, i.e., $(s_2$, $s_3$, $s_6)$ are marked in blue. The key paths either have many sampling points (e.g., $s_1$) or a large length (e.g., $s_7$). $s_6$ is a mask path because it has only a few sampling points, while $s_2$ and $s_3$ are mask paths because they are generated by map matching and do not contain real sampling points from the trajectory.

\subsection{Spatial-temporal-user Joint Embedding} \label{sec:stu_emb}
Path trajectories encompass a wealth of information, including user ID, road segment, and time information. All of this information is important when learning a versatile representation for trajectories. Consequently, we propose a \textit{spatial-temporal-user joint embedding} to encode all the information for model input. Formally, given a segment $v_i$ of a path trajectory, we use Equation~\ref{eq:stu_emb} for encoding:
\begin{equation}
    \label{eq:stu_emb}
    \mathbf{x}_i = \mathbf{h}_i + \mathbf{t}_i +  \mathbf{u}_i,
\end{equation}
where $\mathbf{h}_i$, $\mathbf{t}_i$, and $\mathbf{u}_i$ denote the spatial, time and user encodings of segment $v_i$, which are described next.

\stitle{Spatial Encoding $\mathbf{h}_i$.} The road segments of a city form a complex structure, which can be captured by the road network graph $\mathcal{G}$. We observe that the properties of a segment are affected by itself and its adjacent segments. Therefore, we apply a graph neural network (GNN) to learn embeddings for road segments based on the road network graph $\mathcal{G}$. In particular, GNN computes an embedding for each node in the graph by aggregating the embeddings of its neighbors. We choose the Graph Attention Network (GAT)~\cite{gat}, which adopts an attention mechanism for neighbor aggregation. 

We feed multiple attributes of a segment as the initial input feature of GAT, including the maximum speed limit, average travel time, segment direction, out-degree, in-degree, segment length, and segment type. To be specific, for the maximum speed limit, average travel time, segment direction, and segment length, we apply the min-max normalization. Regarding segment type, we categorize each segment into eight classes, i.e., [\textit{living street}, \textit{motorway}, \textit{primary}, \textit{residential}, \textit{secondary}, \textit{tertiary}, \textit{trunk}, \textit{unclassified}], and we employ one-hot encoding. All these attributes are concatenated to obtain the input features $f_i$ for segment $v_i$. Formally, given the initial feature $f_i$ of segment $v_i$ and road network graph $\mathcal{G}$, GAT computes segment embedding  as $\mathbf{h}_i = \textrm{GAT} (f_i, \mathcal{G})$. We use a 3-layer GAT model to consider 3-hop neighbors for each segment. 

\stitle{Time Encoding $\mathbf{t}_i$.}
We consider two types of time regularities for trajectories. (i) Trajectory patterns vary considerably at different times of a day or a week. For instance, many trajectories move towards office areas on weekday mornings, while evenings usually observe a surge in trajectories heading back home. (ii) The traffic patterns of different road segment types differ at the same time. For example, segments in commercial areas show elevated trajectory volumes during the daytime, whereas segments in residential areas observe increased trajectory data during non-work hours. 

To account for (i), we introduce a time encoding to capture periodicity. Inspired by Time2vec~\cite{time2vec}, given the timestamp $t_i$ of segment $v_i$ in a trajectory, we first transform $t_i$ into a 6-dimension vector $[\textit{hour}, \textit{minute}, \textit{second}, \textit{year}, \textit{month}, \textit{day}]$, then we learn timestamp $t_i$ as a vector $\mathbf{e}_i^t$ of dimension $l/2$ as follows:
\begin{equation}
    \mathbf{e}_i^t = \textrm{FC}_1(t_i) \parallel \text{sin}(\textrm{FC}_2(t_i)),
\end{equation}
where $\textrm{FC}_1(\cdot)$ and $\textrm{FC}_2(\cdot)$ are linear layers to learn time embedding, and $\textrm{sin}(\cdot)$ function helps capture time periodic behaviors. $\parallel$ is a concatenate operation for vectors by channel-dimension. To account for (ii), we design a segment type encoding. In particular, we employ a learnable type encoding matrix $\mathbf{E}^p \in \mathbb{R}^{|T| \times \frac{l}{2}}$, where each row corresponds to a segment type in  $T$ = \{\textit{living street}, \textit{motorway}, \textit{primary}, \textit{residential}, \textit{secondary}, \textrm{\textit{tertiary}}, \textrm{\textit{trunk}}, \textit{unclassified}\}. Given the type of a segment $v_i$, we look up $\mathbf{E}^p$ to obtain its type encoding $\mathbf{e}^p_i$. Then we concatenate the time encoding $\mathbf{e}^t_i$ and segment type encoding $\mathbf{e}^p_i$, and use a fully connected network to interact two encodings $\mathbf{t}_i = FC(\mathbf{e}^t_i \parallel \mathbf{e}^p_i) \in \mathbb{R}^d$, where $\parallel$ means vector concatenation.

\stitle{User Encoding $\mathbf{u}_i$.}
Different drivers have different travel patterns in their trajectories. E.g., an office worker may follow the same route on all weekdays, while taxi drivers try to pick up passengers in crowded areas every day. By combining spatial information and temporal information (joint embedding) with trajectory attributes (dual-objective task learning), the user encoding can capture the specific behaviors of individual users, including travel time, trajectory length, and high-frequency access areas, which benefits tasks such as trajectory similarity computation, travel time estimation, and trajectory classification. To encode user information, we employ a learnable user encoding matrix $\mathbf{E}^u \in \mathbb{R}^{|U| \times l}$, where $|U|$ is the number of users, and each row of $\mathbf{E}^u$ corresponds to a user. 

\subsection{Dual-objective Task Learning}
Here, we present the two learning objectives of RED.

\begin{figure*}[!t]
    \centering
    \includegraphics[width=0.95\linewidth,interpolate=False]{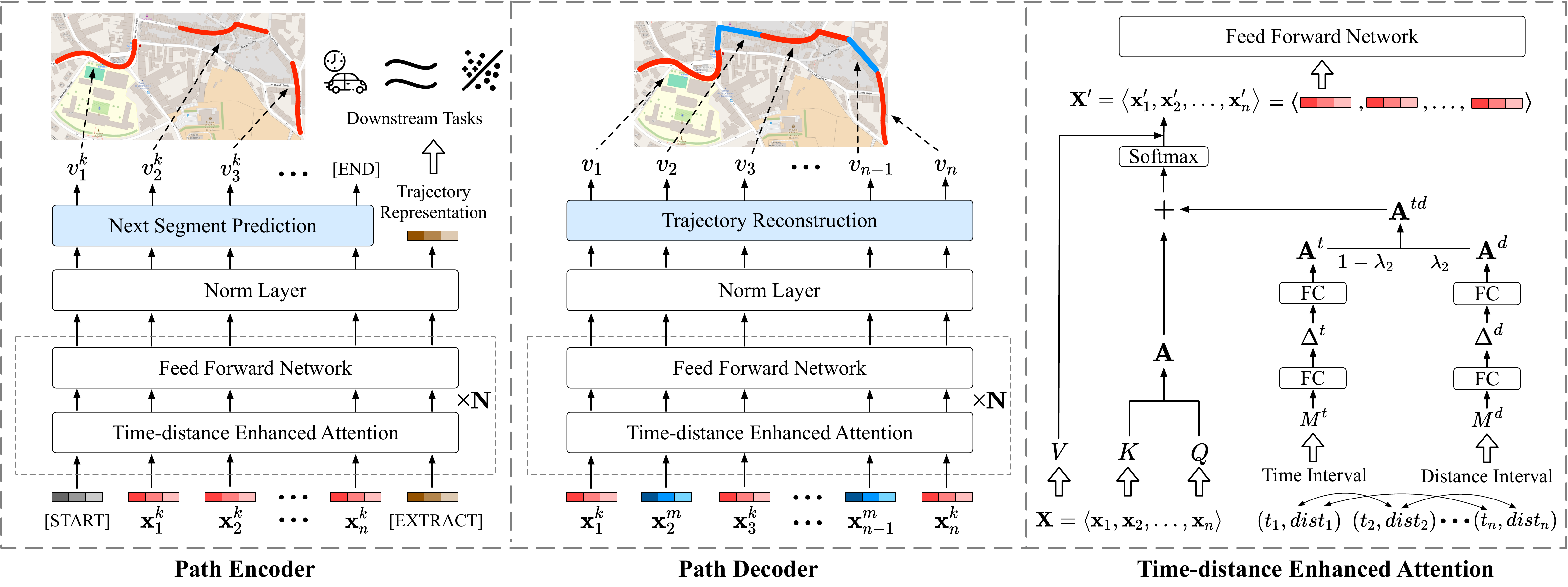}
    \caption{The enhanced Transformer of the encoder and decoder.}
    \label{fig:attention}
\end{figure*} 

\stitle{Path Encoder.}
The path encoder is designed to capture key information of a path trajectory and to predict the next segment for learning local information of the trajectory. Section~\ref{sec:road-aware mask} shows that the key paths of a trajectory carry much more information than the mask paths, suggesting that a trajectory can be described by the key paths. Thus, we require the path encoder to learn information from the key paths. 
Formally, given the key path set of a trajectory $\mathcal{T}^{k} = \langle v_1^k, v_2^k, ..., v_{\vert 
\mathcal{T}^{k} \vert}^k \rangle$, the corresponding key path embedding $\mathbf{X}^{k} = \langle \mathbf{x}_1^k, \mathbf{x}_2^k, ..., \mathbf{x}_{\vert \mathcal{T}^{k} \vert}^k \rangle$ is first generated by the spatial-temporal-user joint embedding, then position encoding~\cite{transformer} is added to $\mathbf{X}^{k}$ to introduce the position information of each segment in the path trajectory. We feed $\mathbf{X}^{k}$ to the Transformer of the encoder and denote $\widehat{\mathbf{X}}^k = \textrm{Transformer}(\mathbf{X}^{k})$ as the output. 

To extract the prediction for the next segment from encoder output, we use a fully connected layer $\mathbf{Y}^k = FC (\widehat{\mathbf{X}}^k)$, where $\mathbf{Y}^k \in \mathbb{R}^{\vert \mathcal{T}^k \vert \times \vert \mathcal{V} \vert}$. Each row of $\mathbf{Y}^k$ is the predicted distribution of a key segment, and each column of $\mathbf{Y}^k$ corresponds to one segment (with $\vert \mathcal{V} \vert$ being the number of all segments). Then we use cross-entropy loss to compute errors between the predicted value and the ground-truth segment ID for the next segment prediction task as follows:
\begin{equation}
    \mathcal{L}^{nsp}_{\mathcal{T}^k} = -\frac{1}{|\mathcal{T}^k|} \sum_{v_i \in \mathcal{T}^k} \log \frac{\exp (\mathbf{Y}_{v_i}^k)}{\sum_{v_j \in \mathcal{V}} \exp (\mathbf{Y}_{v_j}^k)},
\end{equation}
We average the above loss over the trajectories in a mini-batch to obtain the final next segment prediction loss $\mathcal{L}^{nsp}$. 

Note that we predict the next key path in the encoder, so we use \textit{causal self-attention} here to avoid information leakage.

\stitle{Path Decoder.}
The path decoder is designed to restore a raw path trajectory by decoding the embeddings of its segments generated by the encoder. However, the path encoder only encodes the key paths and ignores the mask paths. Although the mask paths carry less information about the trajectory, they are indispensable when restoring a complete trajectory. Thus, we use another Transformer model as the decoder to combine the key paths and the mask paths. We first transform all mask path segments to learnable embedding $\mathbf{X}^{m}$, and combine $\mathbf{X}^{m}$ with the outputs of the path encoder for the key paths. Then we conduct an \textit{unshuffle} operation to restore the mask path embeddings to their original positions in path trajectory $\mathcal{T}$. This process can be formulated as $\mathbf{X}^{\mathcal{T}} = \textrm{Unshuffle}(\textrm{Combine}(\widehat{\mathbf{X}}^k, \mathbf{X}^{m}))$.

The decoder takes $\mathbf{X}^{\mathcal{T}}$ as input and computes hidden embedding output as $\widehat{\mathbf{X}}^{\mathcal{T}} = \textrm{Transformer}(\mathbf{X}^{\mathcal{T}})$. To restore the raw information of trajectory $\mathcal{T}$, we design a trajectory reconstruction task that aims to reconstruct the whole path trajectory from the decoder output $\widehat{\mathbf{X}}^{\mathcal{T}}$. To achieve this, we first use a fully connected layer to obtain the predictions $\mathbf{Y}^{\mathcal{T}} = FC(\widehat{\mathbf{X}}^{\mathcal{T}})$, where $\mathbf{Y}^{\mathcal{T}} \in \mathbb{R}^{\vert \mathcal{T} \vert \times \vert \mathcal{V} \vert}$. Each row of $\mathbf{Y}^{\mathcal{T}}$ is the predicted distribution of a segment of $\mathcal{T}$, and each column of $\mathbf{Y}^{\mathcal{T}}$ corresponds to a segment. Then we use cross-entropy loss for a complete trajectory $\mathcal{T}$ as follows:
\begin{equation}
    \mathcal{L}^{tr}_{\mathcal{T}} = -\frac{1}{|\mathcal{T}|} \sum_{v_i \in \mathcal{T}} \log \frac{\exp (\mathbf{Y}_{v_i}^{\mathcal{T}})}{\sum_{v_j \in \mathcal{V}} \exp (\mathbf{Y}^{\mathcal{T}}_{v_j})},
\end{equation}
We also average the above loss over all trajectories in a mini-batch to obtain the final trajectory reconstruction loss $\mathcal{L}^{tr}$.

Existing TRL methods consider limited supervised information in the training task, e.g., START~\cite{jiang2023start} only reconstructs mask segments of a path trajectory, and Traj2vec~\cite{traj2vec} recovers trajectory points. RED with dual-objective task explores more supervised information and thus can improve the accuracy of trajectory representations. 

\subsection{Enhancing Transformer for Trajectory}
To use Transformer as the model for the path encoder and decoder, two issues need to be addressed. 
First, the next segment prediction task is different from next token prediction of language in subtle ways, and directly using Transformer will encounter errors in the input and output (which we call segment misalignment). Second, the Transformer relies on the self-attention among the token embeddings while trajectories contain spatial-temporal information. To utilize such spatial-temporal information, we need to integrate it into Transformer attention. To fix segment misalignment, we introduce \textit{virtual tokens} for the encoder. To utilize spatial-temporal information of trajectories, we design a \textit{time-distance enhanced attention} to enhance the Transformer for the decoder. Figure~\ref{fig:attention} gives an overview of the enhanced path encoder and decoder.

\subsubsection{Virtual Token} The next segment prediction task of path encoder can encounter segment misalignment. For instance, given path trajectory $\langle v_1, v_2, v_3, v_4, v_5 \rangle$, the key path is $\langle v_1, v_3, v_5 \rangle$ and the mask path is $\langle v_2, v_4 \rangle$. To predict the next key segments, the input of the path encoder is $\langle v_1, v_3 \rangle$, and the expected output of the path encoder is $\langle v_3, v_5 \rangle$. Therefore, the last key segment $v_5$ is not used as input and the first key segment $v_1$ is not used as output. When we combine the output of the encoder with the mask path $\langle v_2, v_4 \rangle$, we will get an incomplete trajectory input $\langle v_2, v_3, v_4, v_5 \rangle$ (i.e., missing $v_1$) for the decoder, which causes incomplete trajectory reconstruction. To solve these problems, we introduce three virtual tokens, i.e., a start token, an end token, and an extract token. 

\stitle{Start Token.}
To complete the first segment in the path encoder output, we introduce a start token [START] at the initiation of all trajectories. However, utilizing [START] to predict the first key segment of a path trajectory cannot consider the topology of the road network. To address this problem, we add [START] as a virtual node in the road network graph $\mathcal{G}$. Specifically, the [START] node is connected to all the nodes of $\mathcal{G}$ with edges, and this augmented graph is also used to generate the spatial encoding.

\stitle{End Token.}
The end token [END] serves as the expected output of the path encoder when taking all key segments of a trajectory as input to predict the next segment.

\stitle{Extract Token.}
To extract the overall information of a trajectory, we introduce an extract token [EXTRACT] at the end of each trajectory. The encoder output embedding for [EXTRACT] is used as the trajectory representation. 

To illustrate the virtual tokens, we take the same path trajectory $\langle v_1, v_2, v_3, v_4, v_5 \rangle$ as an example. Still, the key path is $\langle v_1, v_3, v_5 \rangle$ and the mask path is $\langle v_2, v_4 \rangle$. By adding virtual tokens, the input of the path encoder becomes $\langle\textrm{[START]}, v_1, v_3, v_5, \textrm{[EXTRACT]} \rangle$. The output of the path encoder is to predict $\langle v_1, v_3, v_5, \textrm{[END]} \rangle$, and the output of [EXTRACT] is the trajectory representation. Therefore, the input and output of the path encoder is complete.

\subsubsection{Time-distance Enhanced Attention}
With spatial-temporal-user joint embedding, a path trajectory $\mathcal{T} = \langle \tau_1, \tau_2, ..., \tau_{|\mathcal{T}|} \rangle$ is transformed to embedding $\mathbf{X} = \langle \mathbf{x}_1, \mathbf{x}_2, ..., \mathbf{x}_{|\mathcal{T}|} \rangle  \in \mathbb{R}^{\vert \mathcal{T} \vert \times l} $. Then we can use the vanilla Transformer to model trajectory representation. However, directly applying Equation~\ref{eq:vanilla_sa} neglects the essential spatial-temporal information in trajectories. 
Spatially, a path trajectory consists of continuous road segments, and each segment is related most closely to its adjacent segments. Consequently, two segments that are distant in a path trajectory 
should have small attention scores, while two segments that are close should have large attention scores. Temporally, the sampling timestamps of the segments of a trajectory are consecutive. Two segments with a large sampling time difference should have small attention scores, while the converse is true for two segments with a small sampling time difference. To ensure these properties, we introduce the time-distance enhanced attention to consider the time interval and distance interval between the road segments in a trajectory. In particular, we construct a time-distance attention matrix $\mathbf{A}^{td}$ to incorporate spatial-temporal information to the vanilla Transformer as follows:
\begin{equation}
    \mathbf{X}^\prime = \textrm{softmax} (  \mathbf{A}  + \mathbf{A}^{td} ) V,
    \label{eq:sa}
\end{equation}
where $\mathbf{A}^{td} = (1 - \lambda_2) \mathbf{A}^t + \lambda_2 \mathbf{A}^d$. $\mathbf{A}^t$ is the time interval correlation, $\mathbf{A}^d$ is the distance interval correlation, and $\lambda_2$ is a hyper parameter.

\stitle{Time Interval Correlation.}
Follow by START~\cite{jiang2023start}, consider a path trajectory $\mathcal{T} = \langle \tau_1, \tau_2, ..., \tau_{|\mathcal{T}|} \rangle$, where $\tau_i = (v_i, t_i)$ denotes the segment $v_i$ and timestamp $t_i$ that the trajectory passes through. We first construct a time interval matrix $M^t \in \mathbb{R}^{\vert \mathcal{T} \vert \times \vert 
\mathcal{T} \vert}$, where element $m_{i,j}^t \in M^t$ is the difference $\vert t_i - t_j \vert$ between the sampling timestamps of segments $v_i$ and $v_j$. 
Then, we introduce a transformation $f(m_{i,j}^t) = \frac{1}{\log(e + g_t(m_{i,j}^t))}$ to make larger values smaller to capture correlation based on time interval, where $g_t(m_{i,j}^t) = \frac{m_{i,j}^t}{60}$ maps seconds to minutes to avoid excessive time differences between the segments. In the path encoder, the inputs are key paths of a trajectory, so the time interval matrix is a sub-matrix of $M^t$.

To incorporate the information of $M^t$ into the trajectory representation, we use
a fully connected layers to transform each element $m_{i,j}^t \in M^t$ to a vector in the embedding space as $\delta_{i,j}^t = FC(m_{i,j}^t)$, where the $\delta_{i,j}^t \in \mathbb{R}^{\frac{l}{2}}$. 
Consequently, we obtain the time interval hidden embedding $\Delta^t \in \mathbb{R}^{\vert \mathcal{T} \vert \times \vert \mathcal{T} \vert \times \frac{l}{2}}$ for $\mathcal{T}$.
Then, we use another fully connected layer to transform $\Delta^t$ to time interval correlation matrix $\mathbf{A}^t \in \mathbb{R}^{\vert \mathcal{T} \vert \times \vert \mathcal{T} \vert}$, where each element $\alpha^t_{i,j} \in \mathbf{A}^t$ is a time interval correlation value, and $\alpha^t_{i,j} = FC(\delta_{i,j}^t)$.

\stitle{Distance Interval Correlation.}
Like time interval correlation, we first construct a travel distance interval matrix $M^d \in \mathbb{R}^{\vert \mathcal{T} \vert \times \vert \mathcal{T} \vert}$ for a path trajectory $\mathcal{T}$. In particular, we use the travel length between two segments as $m_{i,j}^d \in M^d$. Then another transformation $f(m_{i,j}^d) = \frac{1}{\log(e + g_d(m_{i,j}^d))}$ is used to ensure that two distant segments have has lower correlation, where $g_d(m_{i,j}^d)) = \frac{m_{i,j}^d}{1000}$ transforms distance from meters to kilometers to avoid excessive values. We also use a fully connected layer to map the transformed distances to the final distance interval correlation matrix $\mathbf{A}^d$.    

\subsection{Discussion}
\stitle{Complexity Analysis.}
The training time of RED encompasses three terms: 
i.e., road-aware masking, graph neural network, and trajectory encoding.
The road-aware masking has time complexity $\mathcal{O}(\vert \mathcal{D} \vert \cdot \vert \mathcal{T} \vert)$, where $\vert \mathcal{T} \vert$ is the average trajectory length and $\vert \mathcal{D} \vert$ is the cardinality of the trajectory dataset. 
The GNN training has complexity $\mathcal{O}(\vert \mathcal{V} \vert \cdot l + \vert \mathcal{E} \vert \cdot l )$, where $l$ is the embedding dimensionality, $\vert \mathcal{V} \vert$ and $\vert \mathcal{E} \vert$ are the numbers of nodes and edges in the road network. 
The complexity of encoding the trajectories is $\mathcal{O}(\vert \mathcal{D} \vert \cdot \vert \mathcal{T}^k \vert^2 \cdot l \cdot L_e + \vert \mathcal{D} \vert \cdot \vert \mathcal{T} \vert^2 \cdot l \cdot L_d)$, where $\vert \mathcal{T}^k \vert$ in the number of key segments in trajectory $\mathcal{T}$ and $\vert \mathcal{T}^k \vert < \vert \mathcal{T} \vert$, $L_e$ and $L_d$ are the numbers of path encoder and path decoder layers. 
To conduct model inference for a trajectory dataset $\mathcal{D}'$, we use the path encoder to compute the vector representations. The complexity of this is $\mathcal{O}(\vert \mathcal{D}' \vert \vert \mathcal{T} \vert^2 \cdot l \cdot L_e)$. The resulting trajectory representations take $\mathcal{O}(l \cdot \vert \mathcal{D}' \vert)$ space.
Using the representations, computing the similarity of two trajectories has time cost  $\mathcal{O}(l)$. In contrast, traditional trajectory similarity algorithms (e.g., DTW, EDR) have cost $\mathcal{O}(\vert \mathcal{D} \vert \cdot \vert \mathcal{T} \vert^2)$ because they use dynamic programming.

\stitle{Limitations.}
RED targets trajectories on the road network (i.e., involving map matching and road segments) and needs to be adjusted for other types of trajectories (e.g., POI trajectories, animal and human flow trajectories). Moreover, RED focuses on trajectory-based tasks (e.g., similarity computation and travel time estimation) and is unsuitable for road network related tasks that do not work at the trajectory level, e.g., road classification and flow estimation.

\begin{table}[!t]
   \centering
   \caption{Statistics of the experiment datasets.}
   \label{tab:dataset}
   \resizebox{0.95\linewidth}{!}{
   \begin{tabular}{l|c|c|c}\toprule
     \textbf{Dataset} & Porto & Rome & Chengdu \\
      \midrule
      \textbf{\#Users} & 439 & 313  & 13,715\\
      \textbf{\#Path trajectories} & 859,986 & 80,203  & 4,413,602 \\
      \textbf{\#Road Segments} & 10,537 & 49,230  & 45,534 \\
      \textbf{Average trajectory length} & 37 & 56  & 26 \\
      \bottomrule
   \end{tabular}
   }
\end{table}

\section{Experimental Evaluation} \label{exp}
We proceed to cover experiments that aim to evaluate and compare RED with the state-of-the-art baselines considering four downstream tasks, trajectory classification, travel time estimation, trajectory similarity computation, and most similar trajectory retrieval on three real-world datasets.

\begin{table*}[!t]
   \LARGE
   \centering
   \caption{Accuracy at travel time estimation. Best-performing existing methods are underlined, and the bottom row reports RED's accuracy improvement over the best existing method.}
   \resizebox{0.71\linewidth}{!}{
   \begin{tabular}{l|ccc|ccc|ccc}
   \toprule
      & \multicolumn{3}{c|}{Porto} & \multicolumn{3}{c|}{Rome} & \multicolumn{3}{c}{Chengdu}
      \\
        & MAE$\downarrow$     & MAPE(\%)$\downarrow$  & RMSE$\downarrow$    
        & MAE$\downarrow$     & MAPE(\%)$\downarrow$  & RMSE$\downarrow$
        & MAE$\downarrow$     & MAPE(\%)$\downarrow$  & RMSE$\downarrow$ \\
        \midrule
		Traj2vec 
	& 1.624  & 20.005  & 3.137    
        & 5.933  & 45.292  & 7.188      
        & 1.397  & 16.833  & 2.114
              \\
		T2vec 
        & 1.608  & 19.906  & 3.043    
        & 5.873  & 44.975 & 7.104    
        & 1.400 & 17.072  & 2.136    
              \\
		Trembr 
        & 1.576  & 19.391  & 2.892  
        & 5.336  & 44.220  & 6.907  
        & 1.374  & 16.635  & 1.937   
              \\
		PIM 
        & 1.684  & 20.137  & 3.205    
        & 5.992  & 45.937  & 7.294
        & 1.622  & 20.897  & 2.796 

              \\
        \midrule
        Transformer
        & 1.844  & 21.325  & 3.790    
        & 6.684  & 47.131  & 9.982
        & 1.707  & 22.458  & 3.031

              \\
        BERT
        & 1.707  & 20.243  & 3.275    
        & 6.114  & 46.142  & 7.358
        & 1.453  & 17.520  & 2.544

              \\
        PIM-TF 
        & 2.018  & 22.456  & 3.898    
        & 6.941  & 48.020  & 8.313  
        & 1.833  & 23.561  & 3.163

              \\
        Toast 
        & 1.801  & 20.952  & 3.334    
        & 6.411  & 46.864  & 7.997  
        & 1.391  & 16.984  & 2.110

              \\
        START 
        & \underline{1.506}  & \underline{18.012}  & \underline{2.799}  
        & \underline{4.342}  & \underline{43.823}  & \underline{6.282}
        & \underline{1.278}  & \underline{15.961}  & \underline{1.889}

        \\
    \midrule
    \textbf{RED} 
        & \textbf{1.468}  & \textbf{16.716}  & \textbf{2.651}     
        & \textbf{3.798}  & \textbf{41.658}  & \textbf{5.358}   
        & \textbf{1.212}  & \textbf{15.648}  & \textbf{1.720}

              \\
		\midrule
		\textbf{Improve.}
        & 2.52\%  & 7.20\%  & 5.29\%    
        & 12.53\%  & 4.94\%  & 14.71\%   
        & 5.16\%  & 1.96\%  & 8.95\%
              \\
		\bottomrule
   \end{tabular}
   }
   \label{tab:tte}
\end{table*}

\begin{table*}[tbp]
   \LARGE
   \centering
   \caption{Accuracy at trajectory classification. Best-performing existing methods are underlined, and the bottom row reports RED's accuracy improvement over the best existing method.}
   \resizebox{0.71\linewidth}{!}{
   \begin{tabular}{l|ccc|ccc|ccc}
   \toprule
      & \multicolumn{3}{c|}{Porto} & \multicolumn{3}{c|}{Rome} & \multicolumn{3}{c}{Chengdu}
      \\
        & Mi-F1$\uparrow$  & Ma-F1$\uparrow$    & Recall@5$\uparrow$  
        & Mi-F1$\uparrow$  & Ma-F1$\uparrow$    & Recall@5$\uparrow$  
        & Accuracy $\uparrow$  & Precision $\uparrow$    & F1$\uparrow$ \\
        \midrule
		Traj2vec 
		& 0.074  & 0.052  & 0.223     
        & 0.069  & 0.047  & 0.185      
        & 0.809  & 0.817  & 0.820  

              \\
		T2vec 
		& 0.083  & 0.062  & 0.238    
        & 0.074  & 0.053  & 0.190   
        & 0.794  & 0.803  & 0.814           
              \\
		Trembr 
		& 0.087  & 0.069  & 0.243    
        & 0.081  & 0.062  & 0.208   
        & 0.811  & 0.836  & 0.848  

              \\
		PIM 
		& 0.069  & 0.041  & 0.207    
        & 0.060  & 0.034  & 0.174 
        & 0.764  & 0.787  & 0.791 

              \\
        \midrule
        Transformer
        & 0.043  & 0.035  & 0.107       
        & 0.041  & 0.034  & 0.098
        & 0.767  & 0.794  & 0.800   

              \\
        BERT
        & 0.077  & 0.054  & 0.229      
        & 0.075  & 0.058  & 0.177
        & 0.810  & 0.837  & 0.852  
              \\
        PIM-TF 
	& 0.035  & 0.028  & 0.083   
        & 0.032  & 0.026  & 0.079   
        & 0.754  & 0.762  & 0.786 

              \\
        Toast 
		& 0.075  & 0.048  & 0.215    
        & 0.070  & 0.044  & 0.206   
        & 0.824 & 0.841  &  0.860

              \\
        START 
		& \underline{0.098}  & \underline{0.076} & \underline{0.251}    
        & \underline{0.095}  & \underline{0.079}  & \underline{0.238} 
        & \underline{0.825}  & \underline{0.849}  & \underline{0.879} 

              \\
		\midrule
		\textbf{RED} 
        & \textbf{0.107}  & \textbf{0.085}  & \textbf{0.262}     
        & \textbf{0.120}  & \textbf{0.106}  & \textbf{0.284}
        & \textbf{0.835}  & \textbf{0.865}  &  \textbf{0.885}
              \\
		\midrule
		\textbf{Improve.}
        & 9.18\%  & 11.84\%    & 4.38\%       
        & 26.32\% &  34.18\%   & 19.33\%  
        & 1.21\%  & 1.88\%  & 0.68\%  
              \\
		\bottomrule
   \end{tabular}
   }
   \label{tab:cls}
\end{table*}

\subsection{Experimental Settings} \label{sec:exp_setting}

\stitle{Datasets.}
We use three real-world datasets: Porto\footnote{https://www.kaggle.com/c/pkdd-15-predict-taxi-service-trajectory-i}, Rome\footnote{https://ieee-dataport.org/open-access/crawdad-Rometaxi}, and Chengdu\footnote{https://www.pkbigdata.com/common/zhzgbCmptDetails.html}. Key statistics are given in Table~\ref{tab:dataset}. We use map data from OpenStreetMap\footnote{https://www.openstreetmap.org}. This data includes road IDs, lengths, road types, speed limits, etc. Trajectory points include latitude, longitude, timestamp, user ID, etc., which are sampled on average every 15, 5, and 30 seconds in Porto, Rome, and Chengdu, respectively. We remove trajectory points that are not within the chosen map boundaries and that are drifting.
We apply the FMM~\cite{fmm} map matching method, which is widely used for road network-based methods~\cite{mao2022JCLRNT,jiang2023start} and is efficient and accurate, to obtain path trajectories. We also remove path trajectories with fewer than 6 segments and set the maximum path trajectory length to 256. The overall proportion of training, validation, and testing data for Porto and Chengdu is set to [0.6, 0.2, 0.2], while for Rome, it is set to [0.8, 0.1, 0.1].

\stitle{Baselines.}  
We compare RED with 9 state-of-the-art TRL methods, including the four \textit{RNN-based methods} \textbf{Traj2vec}~\cite{traj2vec}, \textbf{T2vec}~\cite{t2vec}, \textbf{Trembr}~\cite{trembr}, and \textbf{PIM}~\cite{pim} and the five \textit{Transformer-based methods} \textbf{Transformer}~\cite{transformer}, \textbf{BERT}~\cite{bert}, \textbf{PIM-TF}~\cite{pim}, \textbf{Toast}~\cite{toast}, and 
 \textbf{START}~\cite{jiang2023start}. For trajectory similarity tasks, we further introduce the seven heuristic methods \textbf{SSPD}~\cite{sspd}, \textbf{Hausdorff}~\cite{hausdorff}, \textbf{Fr\'{e}chet}~\cite{frechet}, \textbf{DTW}~\cite{dtw}, \textbf{LCSS}~\cite{lcss}, \textbf{ERP}~\cite{erp}, and \textbf{EDR}~\cite{erp}.

\begin{table*}[!t]
\Large
   \centering
   \caption{Accuracy at trajectory similarity computation. The best-performing baseline is marked with underline, and the bottom row is our accuracy improvement over the best baseline.}
   \resizebox{0.8\linewidth}{!}{
   \begin{tabular}{l|ccc|ccc|ccc|ccc}\toprule
      & \multicolumn{6}{c|}{Porto} & \multicolumn{6}{c}{Rome} \\
      & \multicolumn{3}{c}{Hausdorff distance} &  \multicolumn{3}{c|}{Fr\'{e}chet distance}
      & \multicolumn{3}{c}{Hausdorff distance} &  \multicolumn{3}{c}{Fr\'{e}chet distance}
      \\\cmidrule(lr){2-4}\cmidrule(lr){5-7}\cmidrule(lr){8-10}\cmidrule(lr){11-13}
        & HR@1$\uparrow$ & HR@5$\uparrow$ & HR@10$\uparrow$ 
        & HR@1$\uparrow$ & HR@5$\uparrow$ & HR@10$\uparrow$ 
        & HR@1$\uparrow$ & HR@5$\uparrow$ & HR@10$\uparrow$ 
        & HR@1$\uparrow$ & HR@5$\uparrow$ & HR@10$\uparrow$  \\
        \midrule
		Traj2vec        
            & 0.103 & 0.292  & 0.380   
            & 0.110 & 0.308  & 0.404
            & 0.092 & 0.276  & 0.329  
            & 0.101 & 0.285  & 0.340 \\
		T2vec  
            & 0.129 & 0.330  & 0.431  
            & 0.140 & 0.352  & 0.458
            & 0.121 & 0.304  & 0.361  
            & 0.135 & 0.307  & 0.356 \\
		Trembr    
            & 0.141 & 0.354  & 0.462  
            & 0.164 & 0.377  & 0.476
            & 0.134 & 0.311  & 0.379  
            & 0.147 & 0.320  & 0.384 \\
		PIM 
            & 0.114 & 0.302  & 0.374  
            & 0.129 & 0.321  & 0.399
            & 0.105 & 0.279  & 0.343  
            & 0.111 & 0.286  & 0.350\\
        \midrule
            Transformer
            & 0.083 & 0.260  & 0.351   
            & 0.090 & 0.271  & 0.372
            & 0.079 & 0.244  & 0.302  
            & 0.088 & 0.266  & 0.321 \\
            BERT
            & 0.091 & 0.277  & 0.364   
            & 0.099 & 0.290  & 0.385
            & 0.085 & 0.261  & 0.318  
            & 0.092 & 0.273  & 0.329 \\
            PIM-TF 
            & 0.073 & 0.241  & 0.337  
            & 0.075 & 0.248  & 0.343
            & 0.064 & 0.210  & 0.278  
            & 0.072 & 0.221  & 0.290 \\
            Toast
            & 0.121 & 0.315  & 0.408  
            & 0.137 & 0.320  & 0.417
            & 0.113 & 0.291  & 0.357  
            & 0.128 & 0.302  & 0.366 \\
            START
            & 0.172 & 0.389  & 0.490  
            & 0.185 & 0.427  & 0.532   
            & 0.162 & 0.329  & 0.401 
            & 0.179 & 0.339  & 0.418 \\
            \midrule
		\textbf{RED}   
            & \textbf{0.239} & \textbf{0.501}  & \textbf{0.608}    
            & \textbf{0.246} & \textbf{0.523}  & \textbf{0.638}   
            & \textbf{0.190} & \textbf{0.358}  & \textbf{0.431}  
            & \textbf{0.205} & \textbf{0.387}  & \textbf{0.463} \\
		\midrule
		\textbf{Improve.}    
            & 38.95\% & 28.79\%  & 24.08\%  
            & 32.97\% & 22.48\%  & 19.92\%
            & 17.28\% & 8.81\%  & 7.48\% 
            & 14.53\% & 14.16\%  & 10.77\% \\
		\bottomrule
   \end{tabular}
   }
   \label{tab:approximate_sim}
\end{table*}

\begin{table*}[!t]
\Large
\centering
\caption{Accuracy for most similar trajectory retrieval in terms of mean rank (MR$\downarrow$), where lower values indicate better performance. The best-performing method is highlighted in bold, and the second-best method is underlined.}
\resizebox{0.8\linewidth}{!}{
    \begin{tabular}{l|ccc|ccc|ccc|ccc}
    \toprule
    & \multicolumn{6}{c|}{Porto} 
    & \multicolumn{6}{c}{Rome} 
    \\
      & \multicolumn{3}{c}{$p = 0.1$ on diverse datasize} & \multicolumn{3}{c|}{various $p$ on 100k dataset}
      & \multicolumn{3}{c}{$p = 0.1$ on diverse datasize} & \multicolumn{3}{c}{various $p$ on 40k dataset}
      \\  
      \cmidrule(lr){2-4}\cmidrule(lr){5-7}\cmidrule(lr){8-10}\cmidrule(lr){11-13}

       & 10k & \ \ \ 50k & \ \ \ 100k 
       & $p = 0.2$ & $p = 0.3$ & $p = 0.4$
       & 10k & \ \ \ 20k & \ \ \ 40k
       & $p = 0.2$ & $p = 0.3$ & $p = 0.4$ \\
        \midrule
		Hausdorff    
            & 35.42  & \ \ \ 83.12  & \ \ \ 132.2 
            & 237.3  & 312.0 & 464.1
            & 50.18  & \ \ \ 68.19 & \ \ \ 103.6
            & 236.8  & 262.4  & 340.5  
            \\
		Fr\'{e}chet
            & 45.10  & \ \ \ 88.95 & \ \ \ 159.1 
            & 210.5  & 372.1 & 469.6
            & 187.6  & \ \ \ 212.6 & \ \ \ 437.8
            & 55.55 & 73.62 & 88.70  
            
            \\
		DTW
            & 1.752  & \ \ \ 3.067 & \ \ \ 3.694 
            & 6.298  & 13.01  & 30.60   
            & 1.985  & \ \ \ 2.946 & \ \ \ 4.832  
            & 10.33  & 24.64  & 59.43  
            \\
		LCSS
            & 92.67  & \ \ \ 312.8 & \ \ \ 421.6 
            & 965.9  & 1349  & 2431
            & 125.1  & \ \ \ 147.2 & \ \ \ 350.2
            & 706.7  & 1143  & 1645
            \\
        EDR
            & 146.2  & \ \ \ 447.2 & \ \ \ 667.3 
            & 1090  & 1508 & 2202
            & 176.8  & \ \ \ 232.1 & \ \ \ 455.5
            & 852.3  & 1421  & 1830
            \\
        ERP
            & 32.13  & \ \ \  89.15 & \ \ \ 274.1 
            & 393.0  & 424.5  & 572.6
            & 86.32  & \ \ \ 107.7 & \ \ \ 331.1   
            & 446.7  & 444.0  & 520.8
            \\
        SSPD
            & 2.034  & \ \ \ 3.816 & \ \ \ 4.754 
            & 6.855 & 14.72  & 32.27
            & 3.122  & \ \ \ 5.078 & \ \ \ 8.948   
            & 25.33  & 36.93 & 102.1
            \\
        \midrule
        Traj2vec
        & 1.842  & \ \ \ 2.693  & \ \ \ 2.911 
        & 5.123  & 10.72  & 18.36
        & 1.801  & \ \ \ 2.034 & \ \ \ 2.739   
        & 4.911  & 8.234  & 14.34
        \\
        T2vec
        & 1.740  & \ \ \ 2.351  & \ \ \ 2.400 
        & 5.940  & 10.34  & 17.00
        & 1.661  & \ \ \ 1.839 & \ \ \ 2.177   
        & 4.267  & 7.963  & 15.72
        \\
        Trembr
        & 1.509  & \ \ \ 2.010 & \ \ \ 2.130 
        & 4.536  & 9.681  & 12.74
        & 1.488  & \ \ \ 1.721 & \ \ \ 1.990
        & 3.891  & 6.706  & 10.44
        \\
        PIM
        & 4.331  & \ \ \ 8.937  & \ \ \ 10.98 
        & 40.66  & 67.50  & 161.1
        & 3.912  & \ \ \ 4.771 & \ \ \ 8.324   
        & 36.42  & 63.00  & 92.64

        \\
        Toast
        & 4.997  & \ \ \ 9.310  & \ \ \ 12.43 
        & 44.21  & 84.45  & 132.6
        & 3.949  & \ \ \ 4.801 & \ \ \ 7.330   
       & 29.65  & 48.51  & 78.50
        \\
        START
        & \textbf{1.232}  & \ \ \ \textbf{1.720}  & \ \ \ \textbf{1.847} 
        & \textbf{3.251}  & \textbf{6.241}  & \textbf{8.831}
        & \underline{1.114}  & \ \ \ \underline{1.222} & \ \ \ \underline{1.445}   
        & \underline{3.524}  &  \underline{5.583} & \textbf{7.268}
        \\
	\midrule
        \textbf{RED}
        & \underline{1.420}  & \ \ \ \underline{1.893} & \ \ \ \underline{1.996} 
        & \underline{4.158}  & \underline{7.497}  & \underline{10.77}
        & \textbf{1.096}  & \ \ \ \textbf{1.171} & \ \ \ \textbf{1.343}
        & \textbf{2.955}  & \textbf{4.370} & \underline{7.560}
        \\
	\bottomrule
    \end{tabular}
}
\label{tab:most_sim}
\end{table*}

\stitle{Implementation.} We implement RED using PyTorch 2.0. We use a machine running Ubuntu 20.04 and with an NVIDIA RTX GeForce 4090 GPU. We set the embedding size $l$ to 128 in all methods, the number of GAT layers to 3, and the number of encoder and decoder layers to 6. The attention heads are [8, 16, 1] for GAT, and 8 for the encoder and decoder. The dropout ratio is 0.1. We set $\lambda_1 = 0.1$ for the dual-objective tasks, and $\lambda_2 = 0.5$ for time-distance correlation. We pre-train RED using the AdamW~\cite{adamw} optimizer. The batch sizes are 64, 32, and 32, the training epochs are 10, 30, and 5, and the learning rates are 1e-4 for Porto and Chengdu, 2e-4 for Rome.

\stitle{Downstream Task Settings.}
For all downstream tasks, we only use the encoder of RED and use complete trajectories as input. For trajectory classification and travel-time estimation, we use AdamW~\cite{adamw} to fine-tune the encoder. The batch size is 64, the training epoch is 30, and the learning rate is 1e-4. For trajectory classification on Porto and Rome, we classify trajectories according to user IDs and remove the user encoding to avoid information leakage during fine-tuning. For Chengdu, we classify trajectories based on whether the vehicle is carrying passengers and keep the user encoding during fine-tuning. We add an extra linear layer to get predictions and use cross-entropy as the loss function. For travel time estimation, we remove all time information, including the time encoding as well as the time correlation, except for the departure time, to avoid time information leakage. 
We also add an extra linear layer to predict travel time and use mean square error (MSE) as the loss function. The travel times of all methods are reported in minutes. For trajectory similarity computation, we conduct two types of experiments without fine-tuning, including trajectory similarity computation and most similar trajectory retrieval, and we use the inner-product of trajectory vectors as the similarity score.

\stitle{Performance Metrics.}
For travel time estimation, we report the mean absolute error (MAE), mean absolute percentage error (MAPE), and root mean square error (RMSE). For trajectory multi-classifica\-tion, we report the Micro-F1 (Mi-F1), Macro-F1 (Ma-F1), and Recall@5 values. For the trajectory binary-classification, we report the F1-score, accuracy, and precision. For trajectory similarity tasks, we use Hit Ratio i.e., the HR@1, HR@5, HR@10 values, for trajectory similarity computation and the mean rank (MR) for most similar trajectory retrieval. Ideally, the MR value should be equal to 1.

\subsection{Main Results} \label{sec:exp_main_results}
The main findings for the four downstream tasks are reported at Table~\ref{tab:tte} in travel time estimation, Table~\ref{tab:cls} for trajectory classification, Table~\ref{tab:approximate_sim} for trajectory similarity computation, and Table~\ref{tab:most_sim} for most similar trajectory retrieval. For brevity, we only show results on two datasets for some experiments.

\stitle{Performance at Travel Time Estimation.}
The results for travel time estimation in Table~\ref{tab:tte} show that RED achieves the best performance on all metrics on three real-world datasets. Among the existing methods, Trembr and START perform much better than other methods because these two methods exploit temporal information. On the contrary, disregarding temporal information yields inaccurate predictions, e.g., Toast and PIM. 
Our model not only utilizes temporal correlations in trajectories but also includes a pre-trained time embedding, i.e., Time2vec~\cite{time2vec}, which makes the input of the model carry more temporal features. In addition, we consider the behaviors of the road segment types in different time periods so that the features of road segments and the time are fully integrated.

\begin{table*}[!t]
    \Large
    \centering
    \caption{Accuracy for trajectory classification on the Geolife dataset.}
    \resizebox{0.85\linewidth}{!}{
        \begin{tabular}{l|cccccccc|c}
        \toprule
        Method & SVM~\cite{svm}  & RF~\cite{rf} & TimeLSTM~\cite{time-lstm} & GRU-D~\cite{gru-d} & STGN~\cite{STGCN} & STGRU~\cite{stgru} & TrajODE~\cite{TrajODE} & Trajformer~\cite{Trajformer} & \textbf{RED}    \\
        \midrule
        Accuracy$\uparrow$ & 49.88  & 56.16 & 66.68 & 71.71 & 75.60 & 73.15 & 85.25 & \underline{85.45}  & \textbf{86.89}  \\
        \bottomrule
        \end{tabular}
    }
    \label{tab:geolife_performance}
\end{table*}

\begin{table*}[!t]
\Large
    \centering
    \caption{Average time (in $\upmu$s) to compute the representation for a trajectory.}
    \label{tab:inference}
    \resizebox{0.7\linewidth}{!}{
        \begin{tabular}{l|cccc|ccccc|c}
        \toprule
        & \multicolumn{4}{c|}{RNN-based Method} &  \multicolumn{6}{c}{Transformer-based Method}\\
        \midrule
        Method & Traj2vec  & T2vec & Trembr & PIM & Transformer & BERT & PIM-TF & Toast & START & \textbf{RED}    \\
        \midrule
        Time & 302.25 & 291.90 & 297.33 & 346.00  & 137.80 & 140.83 & 151.93 & 205.92 & 217.80 & 213.30 \\
        \midrule
        Rank & 9 & 7 & 8 & 10  & 1 & 2 & 3 & 4 & 6 & 5 \\
            \bottomrule
        \end{tabular}
    }
\end{table*}

\begin{table}[!t]
\Large
    \centering
    \caption{Average time (in $\upmu$s) to compute one trajectory similarity of RED and different heuristic algorithms.}
    \label{tab:similarity}
    \resizebox{0.99\linewidth}{!}{
        \begin{tabular}{l|ccccccc|c}
        \toprule
        & Hausdorff  & Fr\'{e}chet & DTW & LCSS & EDR & ERP & SSPD & \textbf{RED}    \\
        \midrule
        Porto & 58.48  & 215.55 & 66.87 & 70.68  & 64.62 & 216.52 & 60.30 & \multirow{2}{*}{\textbf{6.14}} \\
        \cmidrule{1-8}
        Rome & 229.85  & 1198.52 & 367.13 & 389.87 & 353.49 &  1194.93 & 232.08 & \\
        \bottomrule
        \end{tabular}
    }
\end{table}

\stitle{Performance at Trajectory Classification.}
The results for trajectory classification are shown in Table~\ref{tab:cls}. RED performs the best because of introducing more comprehensive trajectory information, e.g., spatial information, temporal information, user encoding, and etc. 
Moreover, RED's road-aware masking strategy preserves the crucial information of trajectories, while its graph neural networks and use of static attributes of related road segments as input. The use of comprehensive information can improve the accuracy of trajectory representation and greatly enhance the performance of trajectory classification. 
On the Chengdu dataset, the performance improvement of RED is small because the binary classification task is simple, making it difficult for any method to excel.

\stitle{Performance at Trajectory Similarity Computation.}
We choose two commonly used trajectory similarity heuristics, i.e., Hausdorff distance and Fr\`{e}chet distance, to obtain the most similar trajectory ground truths in the test set. The results of Table~\ref{tab:approximate_sim} show that RED achieves the best performance. RED utilizes comprehensive information on trajectories to improve the expressive of learned trajectory representations and thus supports the high accuracy of trajectory similarity computation.

\stitle{Performance at Most Similar Trajectory Retrieval.}
Following the method~\cite{trajcl}, we create experimental data including a query set $Q$ and a database set $D$ from datasets. Then we downsample the points of each trajectory (if the method uses a road network, we apply detour method~\cite{mao2022JCLRNT}), in the query set $Q$ at ratio $p$ to obtain a sub-trajectory set $Q^\prime$, and then we add $Q^\prime$ to $D$ to obtain the expanded database set $D^\prime$. By construction the trajectories in $Q$ and the trajectories in $Q^\prime$ have high similarity. For each trajectory in query set $Q$, the task aims to find the corresponding sub-trajectory in $D^\prime$. We set $\vert Q \vert = 1,000$, and then set $\vert D \vert = 100,000$ in Porto and $\vert D \vert = 40,000$ in Rome. Table~\ref{tab:most_sim} shows the results.

\textit{Varying the database size $\vert D \vert$.}
We fix the downsampling rate $p$ at 0.1, and vary the database size $\vert D \vert$ from 10,000 to 100,000 in Porto and from 10,000 to 40,000 in Rome. The results show that the average performance of the TRL methods is much higher than that of the heuristic algorithms. 
Because a sub-trajectory has the same semantics as the raw trajectory and TRL methods rely on trajectory semantics, they achieve better performance. Second, the performance of RED is slightly lower than START on Porto and slightly higher than that of START on Rome. The average length of a trajectory in Rome is longer than that of Porto (see Tabel~\ref{tab:dataset}), and through the learning of key paths, our model can capture semantic information of trajectories well in the case of long trajectories. 

\begin{figure}[!t]
    \centering
    \includegraphics[width=\linewidth,interpolate=False]{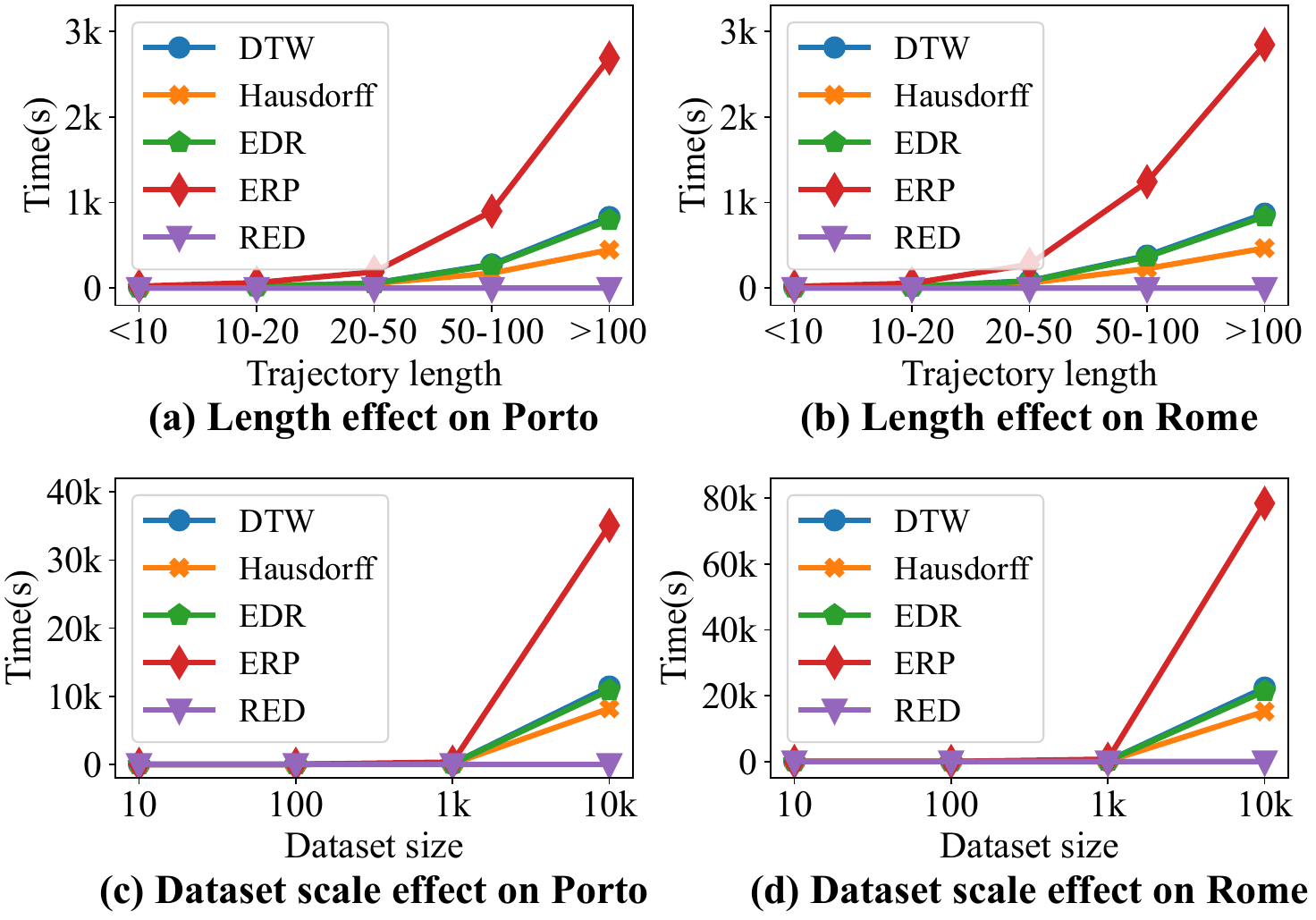}
    \caption{Trajectory similarity computation time (in s) when varying the trajectory length and dataset size.}
    \label{fig:sim_time}
\end{figure}

\textit{Varying downsampling rate $p$.}
We downsample trajectories in $\vert Q \vert$ by different ratios $p \in [0.2, 0.4]$, while fixing $\vert D \vert$ at 100,000 for Porto and 40,000 for Rome. The results show that the average performance of the TRL methods is much higher than that of the heuristic algorithms. The heuristic algorithms perform very poorly at the high downsampling rate $p = 0.4$, while the TRL methods still perform well using their learned semantic information. RED performs slightly below START on Porto and slightly above START on Rome. Because START uses trimming data augmentation during training, it relies on different scales to trim raw trajectories, and this makes START insensitive to downsampling rates.

\stitle{Generalization of RED.}
In addition to the vehicle trajectories, we also apply RED to human mobility trajectories using the Geolife~\cite{geolife-dataset} dataset. When performing trajectory classification~\cite{TrajODE,Trajformer}, Table~\ref{tab:geolife_performance} shows that RED achieves the best accuracy. We further apply RED to sparse trajectories, i.e., POI trajectories, for next POI recommendation~\cite{GETNext,STHGCN} and route recommendation~\cite{SelfTrip,bert-trip}. Here, RED achieves the second-best or the third-best accuracy.

\subsection{Efficiency Study} \label{sec:efficiency}
We study the efficiency on Porto and Rome. In Table~\ref{tab:inference}, we report the average inference time for a trajectory
The inference time is crucial for the efficiency of downstream tasks. The results show that the Transformer-based methods have the shorter inference time than the RNN-based methods. This is because the self-attention in the Transformer can be computed in parallel, while RNNs need to iterate over the paths in a trajectory. The inference time of RED is comparable to those of the Transformer-based methods, which usually perform the best in the accuracy experiments. This suggests that good accuracy comes from tailored model designs instead of high model complexity. 

Computing the similarity between two trajectories is the basic operation in retrieval and clustering. In Table~\ref{tab:similarity}, we report the average time of a trajectory similarity computation. We do not compare RED with other TRL methods because they all transform similarity computation into vector computation and thus have identical computation time when using the same embedding dimensionality. Instead, we compare with heuristic trajectory similarity measures such as DTW, LCSS, etc. The results show that RED is much faster than the heuristic measures. This is because the vector computation of RED has linear complexity while the heuristic measures use dynamic programming, which has quadratic complexity. 

\begin{table}[!t]
   \LARGE
   \centering
   \caption{Accuracy of RED when disabling key designs.}
   \label{tbl:cls}
   \resizebox{\linewidth}{!}{
   \begin{tabular}{l|ccc|ccc}\toprule
      & \multicolumn{3}{c}{Porto} & \multicolumn{3}{c}{Rome}
      \\\cmidrule(lr){2-4}\cmidrule(lr){5-7}  
        & MAPE$\downarrow$  & HR@10$\uparrow$ & Recall@5$\uparrow$ 
        & MAPE$\downarrow$  & HR@10$\uparrow$ & Recall@5$\uparrow$   \\
        & (TTE)  & (TS) & (TC) 
        & (TTE)  & (TS) & (TC)   \\
        \midrule
	  w/o NSP       
            & 18.239  & 0.466  & 0.259     
            & 42.029  & 0.277  & 0.260    \\
	  w/o TE  
            & 18.681  & 0.385  &  0.252
            & 44.835  & 0.336  & 0.274    \\
	w/o VT     
            & 18.537  & 0.394  & 0.257     
            & 43.933  & 0.301  & 0.264    \\
        w/o TD     
            & 17.587 & 0.539  &  0.261    
            & 41.904  & 0.353  & 0.279    \\
		\midrule
		\textbf{RED}       
            & \textbf{16.716}  & \textbf{0.638}  & \textbf{0.262}     
            & \textbf{41.658}  & \textbf{0.463}  & \textbf{0.284}    \\
		\bottomrule
   \end{tabular}
   }
   \label{tab:ablation}
\end{table}

In addition, we explore the impact of trajectory length and dataset size on the efficiency of similarity computation. 
Figure~\ref{fig:sim_time} reports the total time to compute the similarity of trajectory pairs in two datasets. For the trajectory length experiments, we partition trajectories into five sets based on their lengths: [0$\sim$10, 10$\sim$20, 20$\sim$50, 50$\sim$100, 100$\sim$]. Within each set, we randomly sampled 1,000 trajectories and compute the similarity between each pair. Figures~\ref{fig:sim_time}(a) and (b) show that the heuristic algorithms exhibit runtimes that increase notably with longer trajectories. Conversely, RED employs parallel computation on vectors, substantially reducing the computational time. For the dataset size experiments, we vary the dataset size across $[10, 100, 1\textrm{k}, 10\textrm{k}]$ and compute the similarity of each trajectory pair. Figures~\ref{fig:sim_time}(c) and (d) show again that the heuristic algorithms are inefficient. In contrast, parallel computation on vectors is highly efficient and is insensitive to data size fluctuations.

\subsection{Ablation Study and Model Designs} \label{sec:exp_model_design}
\stitle{Ablation Study.}
To gain insight into the contributions of each sub-module in RED, we conduct ablation study on Porto and Rome. Table~\ref{tab:ablation} shows the results. We denote travel time estimation as TTE, trajectory classification as TC, and trajectory similarity as TS. For brevity, we only report on one metric for each task. In trajectory similarity, we show the performance using Fr\'{e}chet distance.
\begin{itemize}[leftmargin=*]
    \item \textbf{w/o NSP}: We remove the next segment prediction task, and keep only the trajectory reconstruction task.
    \item \textbf{w/o time encoding (TE)}: We remove the time encoding and keep only the segment types.
    \item \textbf{w/o virtual token (VT)}: We remove all virtual tokens and the virtual node, and take the prediction of the last position of the encoder as trajectory representation.
    \item \textbf{w/o time-distance (TD)}: We remove the time-distance correlation in self-attention in the encoder and the decoder.
\end{itemize}

Table~\ref{tab:ablation} indicates a pronounced impact at HR@10 performance in trajectory similarity computation. This metric relies on pre-trained trajectory vectors without fine-tuning, necessitating robust generalization of representations. Although reductions in MAPE and Recall@5 performance are also observed, they are comparatively less significant than that of HR@10. 

Additionally, the time encoding, NSP, and virtual token mechanisms play pivotal roles in RED. Time encoding captures crucial temporal information, notably enhancing trajectory representation. 
NSP, tasked with predicting the next segment, is a challenging yet pivotal component. The virtual token addresses data misalignment in the encoder, facilitating accurate prediction of the start position, and thereby preventing incomplete model predictions. 

Furthermore, incorporating the time-distance yields notable enhancements by injecting spatial-temporal information into attention computation, increasing the awareness of neighboring paths within each trajectory path.

\begin{table}[!t]
    \LARGE
   \centering
   \caption{Accuracy of RED with tuned random mask ratio and road-aware masking strategy.}
   \resizebox{\linewidth}{!}{
   \begin{tabular}{l|ccc|ccc}\toprule
      & \multicolumn{3}{c}{Porto} & \multicolumn{3}{c}{Rome}
      \\\cmidrule(lr){2-4}\cmidrule(lr){5-7} 
        & MAPE$\downarrow$  & HR@10$\uparrow$ & Recall@5$\uparrow$    
        & MAPE$\downarrow$  & HR@10$\uparrow$ & Recall@5$\uparrow$   \\
        & (TTE)  & (TS) & (TC) 
        & (TTE)  & (TS) & (TC)   \\
        \midrule
	Ratio 0.1    
            & 17.260  & 0.591  &  0.259    
            & 43.307  & 0.435  & 0.274    \\
	  Ratio 0.3    
            & \underline{17.258}  & \underline{0.602}  & \textbf{0.262}     
            & 43.107  & 0.452      & 0.282    \\
	Ratio 0.5      
            & 17.267  & 0.535  & \textbf{0.262}     
            & \underline{42.331} & \textbf{0.471}  & 0.281    \\
        Ratio 0.7
            & 17.290  & 0.502  & \underline{0.260}     
            & 42.972  & 0.465  & \underline{0.283}    \\
        Ratio 0.9 
            & 17.310  & 0.310  & 0.240 

            & 43.752  & 0.329  & 0.276    \\
		\midrule
		\textbf{RED}     
            & \textbf{16.716}  & \textbf{0.638}  & \textbf{0.262}
            & \textbf{41.658}  & \underline{0.463}  & \textbf{0.284}    \\
		\bottomrule
   \end{tabular}
   }
   \label{tab:mask}
\end{table}

\stitle{Road-aware Masking Strategy.}
The impact of varying random masking ratios, as illustrated in Table~\ref{tab:mask}, is profound. A masking ratio of 0.9 degrades performance severely. Excessive masking impedes RED's ability to extract information from its input, while overly sparse masking oversimplifies the task and prevents the model from learning. The low MAPE variations across different masks are due to the preservation of critical departure and arrival times, pivotal for accurate travel time estimation. 

In contrast, the road-aware masking strategy consistently yields superior performance across metrics. By accounting for the driving patterns in trajectories, this strategy preserves essential road paths and improves training effectiveness. Adaptively adjusting the mask ratio based on trajectory properties ensures a diverse range of mask ratios within the training data. Notably, the automated nature of the road-aware masking strategy obviates the need for manual mask ratio adjustments, streamlining model design and testing.

\begin{figure*}[!t]
    \centering
    \includegraphics[width=\linewidth,interpolate=False]{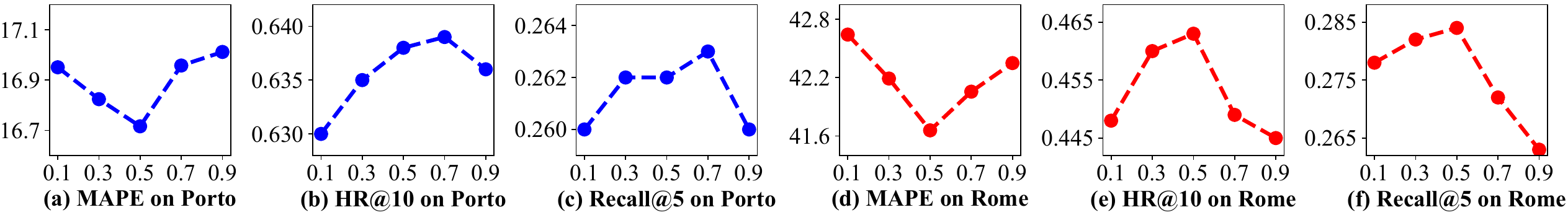}
    \caption{Influence of the time-distance attention ratio (i.e., $\lambda_2$) on the accuracy of RED.}
    \label{fig:td}
\end{figure*}

\stitle{Time-distance Correlation.}
We perform experiments to explore the interplay between time and distance. Specifically, we vary the value of $\lambda_2$ to evaluate the effect of time and distance on the trajectory representations. The results are displayed in Figure~\ref{fig:td}. We show only one metric for each task.  
Figure~\ref{fig:td} shows that the model performance increases and then decreases with the value of $\lambda_2$. Time information produces a higher effect when $\lambda_2 = 0.1$, while distance information produces a higher effect when $\lambda_2 = 0.9$. Both cases show a significant degradation in model performance, which implies that relying on time or distance information alone to influence the attention factor is not effective in learning the local nature of paths. Overall, performance is best when $\lambda_2$ is 0.5.

\balance

\begin{figure}[!t]
    \centering    
    \includegraphics[width=\linewidth,interpolate=False]{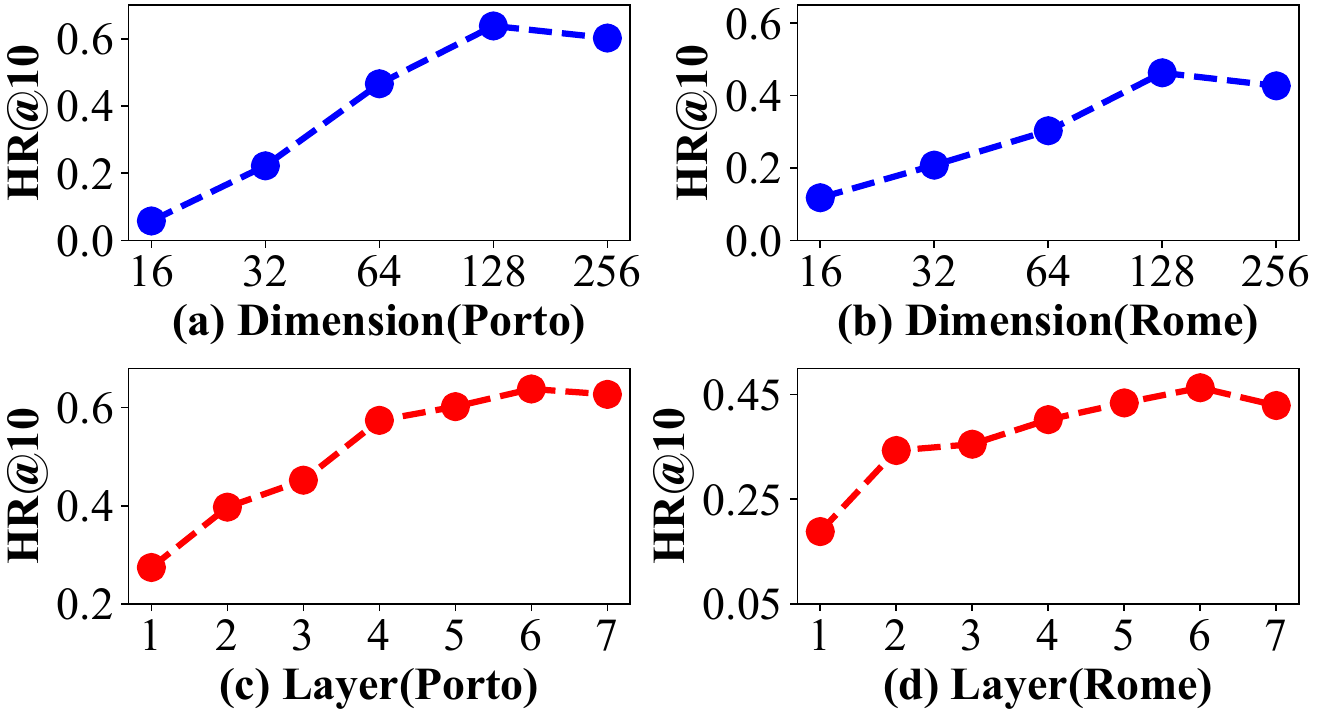}
    \caption{Effect of embedding dimension and model layers.}
    \label{fig:dl}
\end{figure}

\stitle{Performance versus Dimensionality.} 
In this experiment, we vary the dimensionality among 16, 32, 64, 128, and 256, and the remaining hyper parameters are kept at default values. For brevity, we only show the HR@10 values for trajectory similarity computation, exhibit similar performance trends. The results in Figures~\ref{fig:dl} (a) and (b) indicate that the performance of RED increases with the increase in dimensionality and reaches an optimum when $d = 128$ and then starts to decrease. This is because when the dimensionality is small, the representations do not learn enough information, and too large a representation makes the model learn noisy information.

\stitle{Performance versus Layers.}
We perform layer experiments, varying the number of layers among [1, 2, 3, 4, 5, 6, 7], keeping the number of encoder and decoder layers the same and the other hyper parameters are at their default values. We only show the HR@10 values for trajectory similarity computation and other tasks exhibit similar trends. The results in Figures~\ref{fig:dl} (c) and (d) show that the performance of RED increases with increasing numbers of layer and reaches an optimum when layer equals 6 for both datasets and then starts to decrease. When the number of layers is small, the model does not have the ability to extract enough features from the data, while a larger number of layers may incur overfitting.

\section{Related Work}
Trajectory analysis is fundamental to many spatial-temporal analysis tasks and is an important aspect of urban data mining and analytics. Many methods have been proposed to learn
embeddings of trajectories. For example, T2vec~\cite{t2vec}, Traj2SimVec~\cite{traj2simvec}, GTS~\cite{gts}, TrajGAT~\cite{trajgat}, and GRLSTM~\cite{grlstm} learn trajectory representations for similarity computation. Traj2vec~\cite{traj2vec} and E$^2$DTC~\cite{e2dtc} use the seq2seq model for trajectory clustering. Further, GM-VSAE~\cite{gm-vsae} and RL4OASD~\cite{oasd} target anomalous trajectory detection. These methods learn trajectory representations for specific tasks. In contrast, our RED considers generic trajectory representation learning for diverse downstream tasks. Existing generic TRL methods can be classified into RNN-based and Transformer-based methods.

\stitle{RNN-based Methods.}
Trembr~\cite{trembr} uses an RNN-based encoder-decoder and uses the timestamp of each road segment to learn temporal information, but it fails to capture periodic behaviors of different road types in trajectories. PIM~\cite{pim} first uses node2vec~\cite{node2vec} to learn road segment embeddings and then conducts self-supervised RNN-based mutual information maximization to learn trajectory representation. WSCCL~\cite{yang2022wsccl} proposes a weakly supervised contrastive learning method with curriculum learning using LSTMs~\cite{lstm}. HMTRL+~\cite{HMTRL+} uses GRU to model the spatial-temporal correlation of trajectories. These RNN-based methods exhibit long training times, struggle to handle long trajectories and achieve much lower accuracy than Transformer-based methods.

\stitle{Transformer-based Methods.}
Since trajectories are sequential structures, that fit very well with the Transformer, transformer-based methods have become the main focus of the TRL research. Toast~\cite{toast} proposes a traffic context aware skip-gram~\cite{skip-gram} module and a trajectory-enhanced Transformer module to learn trajectory representation. JCLRNT~\cite{mao2022JCLRNT} offers domain-specific augmentations for road-road contrast and trajectory-trajectory contrast to incorporate valuable inter-relations. START~\cite{jiang2023start} incorporates temporal regularities and travel semantics into generic TRL by random masking strategy. START also features data augmentation techniques involving trajectory trimming, temporal shifting, and contrastive learning. LightPath~\cite{lightpath} uses different mask ratios to augment trajectories and then applies a teacher-student contrastive model to learn trajectory representations. However, these approaches either ignore spatial-temporal information (e.g., LightPath, Toast) or rely heavily on contrastive learning for data augmentation (e.g., JCLRNT, START, LightPath). In particular, the same data augmentation technique generally does not work well across different datasets, and also doubles the training dataset and increases the training time. In contrast, RED considers the comprehensive information of the trajectories and does not require data augmentation technologies.

\section{Conclusion}
We propose RED, a framework that utilizes self-supervised learning for trajectory representation learning. Motivated by the limitations of existing studies, RED aims to utilize comprehensive information in trajectory, including road, user, spatial, temporal, travel, and movement. For such purpose, RED features three key designs: road-aware masking strategy, spatial-temporal-user joint embedding, and dual-objective task learning. Experimental results show that RED consistently outperforms existing methods in terms of accuracy across different datasets and downstream tasks, thus advancing the state-of-the-art in trajectory representation learning.

\section{Acknowledgement}
This paper was supported by the National Key R\&D Program of China 2023YFC3305600, 2024YFE0111800, and NSFC U22B2037, U21B2046, and 62032001.

\newpage
\bibliographystyle{ACM-Reference-Format}
\bibliography{Ref_vldb}


\begin{thebibliography}{52}


\ifx \showCODEN    \undefined \def \showCODEN     #1{\unskip}     \fi
\ifx \showDOI      \undefined \def \showDOI       #1{#1}\fi
\ifx \showISBNx    \undefined \def \showISBNx     #1{\unskip}     \fi
\ifx \showISBNxiii \undefined \def \showISBNxiii  #1{\unskip}     \fi
\ifx \showISSN     \undefined \def \showISSN      #1{\unskip}     \fi
\ifx \showLCCN     \undefined \def \showLCCN      #1{\unskip}     \fi
\ifx \shownote     \undefined \def \shownote      #1{#1}          \fi
\ifx \showarticletitle \undefined \def \showarticletitle #1{#1}   \fi
\ifx \showURL      \undefined \def \showURL       {\relax}        \fi
\providecommand\bibfield[2]{#2}
\providecommand\bibinfo[2]{#2}
\providecommand\natexlab[1]{#1}
\providecommand\showeprint[2][]{arXiv:#2}

\bibitem[Alt(2009)]%
        {hausdorff}
\bibfield{author}{\bibinfo{person}{Helmut Alt}.} \bibinfo{year}{2009}\natexlab{}.
\newblock \showarticletitle{The Computational Geometry of Comparing Shapes}.
\newblock \bibinfo{journal}{\emph{Efficient Algorithms, Essays Dedicated to Kurt Mehlhorn on the Occasion of His 60th Birthday}}  \bibinfo{volume}{5760} (\bibinfo{year}{2009}), \bibinfo{pages}{235--248}.
\newblock


\bibitem[Alt and Godau(1995)]%
        {frechet}
\bibfield{author}{\bibinfo{person}{Helmut Alt} {and} \bibinfo{person}{Michael Godau}.} \bibinfo{year}{1995}\natexlab{}.
\newblock \showarticletitle{Computing the Fr{\'{e}}chet distance between two polygonal curves}.
\newblock \bibinfo{journal}{\emph{International Journal of Computational Geometry \& Applications}}  \bibinfo{volume}{5} (\bibinfo{year}{1995}), \bibinfo{pages}{75--91}.
\newblock


\bibitem[Berndt and Clifford(1994)]%
        {dtw}
\bibfield{author}{\bibinfo{person}{Donald~J. Berndt} {and} \bibinfo{person}{James Clifford}.} \bibinfo{year}{1994}\natexlab{}.
\newblock \showarticletitle{Using Dynamic Time Warping to Find Patterns in Time Series}. In \bibinfo{booktitle}{\emph{AAAI}}. \bibinfo{pages}{359--370}.
\newblock


\bibitem[Besse et~al\mbox{.}(2015)]%
        {sspd}
\bibfield{author}{\bibinfo{person}{Philippe~C. Besse}, \bibinfo{person}{Brendan Guillouet}, \bibinfo{person}{Jean{-}Michel Loubes}, {and} \bibinfo{person}{Fran{\c{c}}ois Royer}.} \bibinfo{year}{2015}\natexlab{}.
\newblock \showarticletitle{Review and perspective for distance based trajectory clustering}. In \bibinfo{booktitle}{\emph{arXiv preprint}}. \bibinfo{pages}{http://arxiv.org/pdf/1508.04904}.
\newblock


\bibitem[Breiman(2001)]%
        {rf}
\bibfield{author}{\bibinfo{person}{Leo Breiman}.} \bibinfo{year}{2001}\natexlab{}.
\newblock \showarticletitle{Random Forests}.
\newblock \bibinfo{journal}{\emph{Machine learning}} \bibinfo{volume}{45}, \bibinfo{number}{1} (\bibinfo{year}{2001}), \bibinfo{pages}{5--32}.
\newblock


\bibitem[Chang et~al\mbox{.}(2023)]%
        {trajcl}
\bibfield{author}{\bibinfo{person}{Yanchuan Chang}, \bibinfo{person}{Jianzhong Qi}, \bibinfo{person}{Yuxuan Liang}, {and} \bibinfo{person}{Egemen Tanin}.} \bibinfo{year}{2023}\natexlab{}.
\newblock \showarticletitle{Contrastive Trajectory Similarity Learning with Dual-Feature Attention}. In \bibinfo{booktitle}{\emph{ICDE}}. \bibinfo{pages}{2933--2945}.
\newblock


\bibitem[Che et~al\mbox{.}(2018)]%
        {gru-d}
\bibfield{author}{\bibinfo{person}{Zhengping Che}, \bibinfo{person}{Sanjay Purushotham}, \bibinfo{person}{Kyunghyun Cho}, \bibinfo{person}{David Sontag}, {and} \bibinfo{person}{Yan Liu}.} \bibinfo{year}{2018}\natexlab{}.
\newblock \showarticletitle{Recurrent neural networks for multivariate time series with missing values}.
\newblock \bibinfo{journal}{\emph{Scientific reports}} \bibinfo{volume}{8}, \bibinfo{number}{1} (\bibinfo{year}{2018}), \bibinfo{pages}{6085}.
\newblock


\bibitem[Chen and Ng(2004)]%
        {erp}
\bibfield{author}{\bibinfo{person}{Lei Chen} {and} \bibinfo{person}{Raymond~T. Ng}.} \bibinfo{year}{2004}\natexlab{}.
\newblock \showarticletitle{On The Marriage of Lp-norms and Edit Distance}. In \bibinfo{booktitle}{\emph{VLDB}}. \bibinfo{pages}{792--803}.
\newblock


\bibitem[Chen et~al\mbox{.}(2005)]%
        {edr}
\bibfield{author}{\bibinfo{person}{Lei Chen}, \bibinfo{person}{M.~Tamer {\"{O}}zsu}, {and} \bibinfo{person}{Vincent Oria}.} \bibinfo{year}{2005}\natexlab{}.
\newblock \showarticletitle{Robust and Fast Similarity Search for Moving Object Trajectories}. In \bibinfo{booktitle}{\emph{SIGMOD}}. \bibinfo{pages}{491--502}.
\newblock


\bibitem[Chen et~al\mbox{.}(2021)]%
        {toast}
\bibfield{author}{\bibinfo{person}{Yile Chen}, \bibinfo{person}{Xiucheng Li}, \bibinfo{person}{Gao Cong}, \bibinfo{person}{Zhifeng Bao}, \bibinfo{person}{Cheng Long}, \bibinfo{person}{Yiding Liu}, \bibinfo{person}{Arun~Kumar Chandran}, {and} \bibinfo{person}{Richard Ellison}.} \bibinfo{year}{2021}\natexlab{}.
\newblock \showarticletitle{Robust Road Network Representation Learning: When Traffic Patterns Meet Traveling Semantics}. In \bibinfo{booktitle}{\emph{CIKM}}. \bibinfo{pages}{211--220}.
\newblock


\bibitem[Devlin et~al\mbox{.}(2019)]%
        {bert}
\bibfield{author}{\bibinfo{person}{Jacob Devlin}, \bibinfo{person}{Ming{-}Wei Chang}, \bibinfo{person}{Kenton Lee}, {and} \bibinfo{person}{Kristina Toutanova}.} \bibinfo{year}{2019}\natexlab{}.
\newblock \showarticletitle{{BERT:} Pre-training of Deep Bidirectional Transformers for Language Understanding}. In \bibinfo{booktitle}{\emph{NAACL}}. \bibinfo{pages}{4171--4186}.
\newblock


\bibitem[Fang et~al\mbox{.}(2021)]%
        {e2dtc}
\bibfield{author}{\bibinfo{person}{Ziquan Fang}, \bibinfo{person}{Yuntao Du}, \bibinfo{person}{Lu Chen}, \bibinfo{person}{Yujia Hu}, \bibinfo{person}{Yunjun Gao}, {and} \bibinfo{person}{Gang Chen}.} \bibinfo{year}{2021}\natexlab{}.
\newblock \showarticletitle{E\({}^{\mbox{2}}\)DTC: An End to End Deep Trajectory Clustering Framework via Self-Training}. In \bibinfo{booktitle}{\emph{ICDE}}. \bibinfo{pages}{696--707}.
\newblock


\bibitem[Fu and Lee(2020)]%
        {trembr}
\bibfield{author}{\bibinfo{person}{Tao{-}Yang Fu} {and} \bibinfo{person}{Wang{-}Chien Lee}.} \bibinfo{year}{2020}\natexlab{}.
\newblock \showarticletitle{Trembr: Exploring road networks for trajectory representation learning}.
\newblock \bibinfo{journal}{\emph{ACM Transactions on Intelligent Systems and Technology}} \bibinfo{volume}{11}, \bibinfo{number}{1} (\bibinfo{year}{2020}), \bibinfo{pages}{1--25}.
\newblock


\bibitem[Gao et~al\mbox{.}(2022)]%
        {SelfTrip}
\bibfield{author}{\bibinfo{person}{Qiang Gao}, \bibinfo{person}{Wei Wang}, \bibinfo{person}{Kunpeng Zhang}, \bibinfo{person}{Xin Yang}, \bibinfo{person}{Congcong Miao}, {and} \bibinfo{person}{Tianrui Li}.} \bibinfo{year}{2022}\natexlab{}.
\newblock \showarticletitle{Self-supervised representation learning for trip recommendation}.
\newblock \bibinfo{journal}{\emph{Knowledge-Based Systems}}  \bibinfo{volume}{247} (\bibinfo{year}{2022}), \bibinfo{pages}{108791}.
\newblock


\bibitem[Grover and Leskovec(2016)]%
        {node2vec}
\bibfield{author}{\bibinfo{person}{Aditya Grover} {and} \bibinfo{person}{Jure Leskovec}.} \bibinfo{year}{2016}\natexlab{}.
\newblock \showarticletitle{node2vec: Scalable Feature Learning for Networks}. In \bibinfo{booktitle}{\emph{KDD}}. \bibinfo{pages}{855--864}.
\newblock


\bibitem[Han et~al\mbox{.}(2021)]%
        {gts}
\bibfield{author}{\bibinfo{person}{Peng Han}, \bibinfo{person}{Jin Wang}, \bibinfo{person}{Di Yao}, \bibinfo{person}{Shuo Shang}, {and} \bibinfo{person}{Xiangliang Zhang}.} \bibinfo{year}{2021}\natexlab{}.
\newblock \showarticletitle{A Graph-based Approach for Trajectory Similarity Computation in Spatial Networks}. In \bibinfo{booktitle}{\emph{KDD}}. \bibinfo{pages}{556--564}.
\newblock


\bibitem[He et~al\mbox{.}(2022)]%
        {mae}
\bibfield{author}{\bibinfo{person}{Kaiming He}, \bibinfo{person}{Xinlei Chen}, \bibinfo{person}{Saining Xie}, \bibinfo{person}{Yanghao Li}, \bibinfo{person}{Piotr Doll{\'{a}}r}, {and} \bibinfo{person}{Ross~B. Girshick}.} \bibinfo{year}{2022}\natexlab{}.
\newblock \showarticletitle{Masked Autoencoders Are Scalable Vision Learners}. In \bibinfo{booktitle}{\emph{CVPR}}. \bibinfo{pages}{15979--15988}.
\newblock


\bibitem[Hearst et~al\mbox{.}(1998)]%
        {svm}
\bibfield{author}{\bibinfo{person}{M.A. Hearst}, \bibinfo{person}{S.T. Dumais}, \bibinfo{person}{E. Osuna}, \bibinfo{person}{J. Platt}, {and} \bibinfo{person}{B. Scholkopf}.} \bibinfo{year}{1998}\natexlab{}.
\newblock \showarticletitle{Support vector machines}.
\newblock \bibinfo{journal}{\emph{IEEE Intelligent Systems and their Applications}} \bibinfo{volume}{13}, \bibinfo{number}{4} (\bibinfo{year}{1998}), \bibinfo{pages}{18--28}.
\newblock


\bibitem[Hochreiter and Schmidhuber(1997)]%
        {lstm}
\bibfield{author}{\bibinfo{person}{Sepp Hochreiter} {and} \bibinfo{person}{J{\"{u}}rgen Schmidhuber}.} \bibinfo{year}{1997}\natexlab{}.
\newblock \showarticletitle{Long Short-Term Memory}.
\newblock \bibinfo{journal}{\emph{Neural Computation}} \bibinfo{volume}{9}, \bibinfo{number}{8} (\bibinfo{year}{1997}), \bibinfo{pages}{1735--1780}.
\newblock


\bibitem[Jiang et~al\mbox{.}(2023)]%
        {jiang2023start}
\bibfield{author}{\bibinfo{person}{Jiawei Jiang}, \bibinfo{person}{Dayan Pan}, \bibinfo{person}{Houxing Ren}, \bibinfo{person}{Xiaohan Jiang}, \bibinfo{person}{Chao Li}, {and} \bibinfo{person}{Jingyuan Wang}.} \bibinfo{year}{2023}\natexlab{}.
\newblock \showarticletitle{Self-supervised Trajectory Representation Learning with Temporal Regularities and Travel Semantics}. In \bibinfo{booktitle}{\emph{ICDE}}. \bibinfo{pages}{843--855}.
\newblock


\bibitem[Kazemi et~al\mbox{.}(2019)]%
        {time2vec}
\bibfield{author}{\bibinfo{person}{Seyed~Mehran Kazemi}, \bibinfo{person}{Rishab Goel}, \bibinfo{person}{Sepehr Eghbali}, \bibinfo{person}{Janahan Ramanan}, \bibinfo{person}{Jaspreet Sahota}, \bibinfo{person}{Sanjay Thakur}, \bibinfo{person}{Stella Wu}, \bibinfo{person}{Cathal Smyth}, \bibinfo{person}{Pascal Poupart}, {and} \bibinfo{person}{Marcus~A. Brubaker}.} \bibinfo{year}{2019}\natexlab{}.
\newblock \showarticletitle{Time2vec: Learning a vector representation of time}. In \bibinfo{booktitle}{\emph{arXiv preprint}}. \bibinfo{pages}{https://arxiv.org/pdf/1907.05321}.
\newblock


\bibitem[Kuo et~al\mbox{.}(2023)]%
        {bert-trip}
\bibfield{author}{\bibinfo{person}{Ai{-}Te Kuo}, \bibinfo{person}{Haiquan Chen}, {and} \bibinfo{person}{Wei{-}Shinn Ku}.} \bibinfo{year}{2023}\natexlab{}.
\newblock \showarticletitle{BERT-Trip: Effective and Scalable Trip Representation using Attentive Contrast Learning}. In \bibinfo{booktitle}{\emph{ICDE}}. \bibinfo{pages}{612--623}.
\newblock


\bibitem[Li et~al\mbox{.}(2018)]%
        {t2vec}
\bibfield{author}{\bibinfo{person}{Xiucheng Li}, \bibinfo{person}{Kaiqi Zhao}, \bibinfo{person}{Gao Cong}, \bibinfo{person}{Christian~S. Jensen}, {and} \bibinfo{person}{Wei Wei}.} \bibinfo{year}{2018}\natexlab{}.
\newblock \showarticletitle{Deep Representation Learning for Trajectory Similarity Computation}. In \bibinfo{booktitle}{\emph{ICDE}}. \bibinfo{pages}{617--628}.
\newblock


\bibitem[Liang et~al\mbox{.}(2022)]%
        {Trajformer}
\bibfield{author}{\bibinfo{person}{Yuxuan Liang}, \bibinfo{person}{Kun Ouyang}, \bibinfo{person}{Yiwei Wang}, \bibinfo{person}{Xu Liu}, \bibinfo{person}{Hongyang Chen}, \bibinfo{person}{Junbo Zhang}, \bibinfo{person}{Yu Zheng}, {and} \bibinfo{person}{Roger Zimmermann}.} \bibinfo{year}{2022}\natexlab{}.
\newblock \showarticletitle{TrajFormer: Efficient Trajectory Classification with Transformers}. In \bibinfo{booktitle}{\emph{CIKM}}. \bibinfo{pages}{1229--1237}.
\newblock


\bibitem[Liang et~al\mbox{.}(2021)]%
        {TrajODE}
\bibfield{author}{\bibinfo{person}{Yuxuan Liang}, \bibinfo{person}{Kun Ouyang}, \bibinfo{person}{Hanshu Yan}, \bibinfo{person}{Yiwei Wang}, \bibinfo{person}{Zekun Tong}, {and} \bibinfo{person}{Roger Zimmermann}.} \bibinfo{year}{2021}\natexlab{}.
\newblock \showarticletitle{Modeling Trajectories with Neural Ordinary Differential Equations}. In \bibinfo{booktitle}{\emph{IJCAI}}. \bibinfo{pages}{1498--1504}.
\newblock


\bibitem[Liu et~al\mbox{.}(2023a)]%
        {HMTRL+}
\bibfield{author}{\bibinfo{person}{Hao Liu}, \bibinfo{person}{Jindong Han}, \bibinfo{person}{Yanjie Fu}, \bibinfo{person}{Yanyan Li}, \bibinfo{person}{Kai Chen}, {and} \bibinfo{person}{Hui Xiong}.} \bibinfo{year}{2023}\natexlab{a}.
\newblock \showarticletitle{Unified route representation learning for multi-modal transportation recommendation with spatiotemporal pre-training}.
\newblock \bibinfo{journal}{\emph{The VLDB Journal}} \bibinfo{volume}{32}, \bibinfo{number}{2} (\bibinfo{year}{2023}), \bibinfo{pages}{325--342}.
\newblock


\bibitem[Liu et~al\mbox{.}(2023b)]%
        {tte1}
\bibfield{author}{\bibinfo{person}{Hao Liu}, \bibinfo{person}{Wenzhao Jiang}, \bibinfo{person}{Shui Liu}, {and} \bibinfo{person}{Xi Chen}.} \bibinfo{year}{2023}\natexlab{b}.
\newblock \showarticletitle{Uncertainty-Aware Probabilistic Travel Time Prediction for On-Demand Ride-Hailing at DiDi}. In \bibinfo{booktitle}{\emph{KDD}}. \bibinfo{pages}{4516--4526}.
\newblock


\bibitem[Liu et~al\mbox{.}(2019)]%
        {stgru}
\bibfield{author}{\bibinfo{person}{Hongbin Liu}, \bibinfo{person}{Hao Wu}, \bibinfo{person}{Weiwei Sun}, {and} \bibinfo{person}{Ickjai Lee}.} \bibinfo{year}{2019}\natexlab{}.
\newblock \showarticletitle{Spatio-Temporal {GRU} for Trajectory Classification}. In \bibinfo{booktitle}{\emph{ICDM}}. \bibinfo{pages}{1228--1233}.
\newblock


\bibitem[Liu et~al\mbox{.}(2020)]%
        {gm-vsae}
\bibfield{author}{\bibinfo{person}{Yiding Liu}, \bibinfo{person}{Kaiqi Zhao}, \bibinfo{person}{Gao Cong}, {and} \bibinfo{person}{Zhifeng Bao}.} \bibinfo{year}{2020}\natexlab{}.
\newblock \showarticletitle{Online anomalous trajectory detection with deep generative sequence modeling}. In \bibinfo{booktitle}{\emph{ICDE}}. \bibinfo{pages}{949--960}.
\newblock


\bibitem[Loshchilov and Hutter(2019)]%
        {adamw}
\bibfield{author}{\bibinfo{person}{Ilya Loshchilov} {and} \bibinfo{person}{Frank Hutter}.} \bibinfo{year}{2019}\natexlab{}.
\newblock \showarticletitle{Decoupled Weight Decay Regularization}. In \bibinfo{booktitle}{\emph{ICLR}}. \bibinfo{pages}{https://arxiv.org/pdf/1711.05101}.
\newblock


\bibitem[Mao et~al\mbox{.}(2022)]%
        {mao2022JCLRNT}
\bibfield{author}{\bibinfo{person}{Zhenyu Mao}, \bibinfo{person}{Ziyue Li}, \bibinfo{person}{Dedong Li}, \bibinfo{person}{Lei Bai}, {and} \bibinfo{person}{Rui Zhao}.} \bibinfo{year}{2022}\natexlab{}.
\newblock \showarticletitle{Jointly contrastive representation learning on road network and trajectory}. In \bibinfo{booktitle}{\emph{CIKM}}. \bibinfo{pages}{1501--1510}.
\newblock


\bibitem[Mikolov et~al\mbox{.}(2013)]%
        {skip-gram}
\bibfield{author}{\bibinfo{person}{Tom{\'{a}}s Mikolov}, \bibinfo{person}{Kai Chen}, \bibinfo{person}{Greg Corrado}, {and} \bibinfo{person}{Jeffrey Dean}.} \bibinfo{year}{2013}\natexlab{}.
\newblock \showarticletitle{Efficient Estimation of Word Representations in Vector Space}. In \bibinfo{booktitle}{\emph{ICLR}}. \bibinfo{pages}{https://arxiv.org/pdf/1301.3781}.
\newblock


\bibitem[Pedersen et~al\mbox{.}(2020)]%
        {pedersen2020urban_planning}
\bibfield{author}{\bibinfo{person}{Simon~Aagaard Pedersen}, \bibinfo{person}{Bin Yang}, {and} \bibinfo{person}{Christian~S. Jensen}.} \bibinfo{year}{2020}\natexlab{}.
\newblock \showarticletitle{Fast stochastic routing under time-varying uncertainty}.
\newblock \bibinfo{journal}{\emph{The VLDB Journal}} \bibinfo{volume}{29}, \bibinfo{number}{4} (\bibinfo{year}{2020}), \bibinfo{pages}{819--839}.
\newblock


\bibitem[Vaswani et~al\mbox{.}(2017)]%
        {transformer}
\bibfield{author}{\bibinfo{person}{Ashish Vaswani}, \bibinfo{person}{Noam Shazeer}, \bibinfo{person}{Niki Parmar}, \bibinfo{person}{Jakob Uszkoreit}, \bibinfo{person}{Llion Jones}, \bibinfo{person}{Aidan~N. Gomez}, \bibinfo{person}{Lukasz Kaiser}, {and} \bibinfo{person}{Illia Polosukhin}.} \bibinfo{year}{2017}\natexlab{}.
\newblock \showarticletitle{Attention is All you Need}. In \bibinfo{booktitle}{\emph{NIPS}}. \bibinfo{pages}{5998--6008}.
\newblock


\bibitem[Velickovic et~al\mbox{.}(2018)]%
        {gat}
\bibfield{author}{\bibinfo{person}{Petar Velickovic}, \bibinfo{person}{Guillem Cucurull}, \bibinfo{person}{Arantxa Casanova}, \bibinfo{person}{Adriana Romero}, \bibinfo{person}{Pietro Li{\`{o}}}, {and} \bibinfo{person}{Yoshua Bengio}.} \bibinfo{year}{2018}\natexlab{}.
\newblock \showarticletitle{Graph Attention Networks}. In \bibinfo{booktitle}{\emph{ICLR}}. \bibinfo{pages}{https://arxiv.org/pdf/1710.10903}.
\newblock


\bibitem[Vlachos et~al\mbox{.}(2002)]%
        {lcss}
\bibfield{author}{\bibinfo{person}{Michail Vlachos}, \bibinfo{person}{Dimitrios Gunopulos}, {and} \bibinfo{person}{George Kollios}.} \bibinfo{year}{2002}\natexlab{}.
\newblock \showarticletitle{Discovering Similar Multidimensional Trajectories}. In \bibinfo{booktitle}{\emph{ICDE}}. \bibinfo{pages}{673--684}.
\newblock


\bibitem[Yan et~al\mbox{.}(2023)]%
        {STHGCN}
\bibfield{author}{\bibinfo{person}{Xiaodong Yan}, \bibinfo{person}{Tengwei Song}, \bibinfo{person}{Yifeng Jiao}, \bibinfo{person}{Jianshan He}, \bibinfo{person}{Jiaotuan Wang}, \bibinfo{person}{Ruopeng Li}, {and} \bibinfo{person}{Wei Chu}.} \bibinfo{year}{2023}\natexlab{}.
\newblock \showarticletitle{Spatio-Temporal Hypergraph Learning for Next {POI} Recommendation}. In \bibinfo{booktitle}{\emph{SIGIR}}. \bibinfo{pages}{403--412}.
\newblock


\bibitem[Yang et~al\mbox{.}(2021a)]%
        {yang2021transportation_optimization}
\bibfield{author}{\bibinfo{person}{Chengcheng Yang}, \bibinfo{person}{Lisi Chen}, \bibinfo{person}{Hao Wang}, {and} \bibinfo{person}{Shuo Shang}.} \bibinfo{year}{2021}\natexlab{a}.
\newblock \showarticletitle{Towards Efficient Selection of Activity Trajectories based on Diversity and Coverage}. In \bibinfo{booktitle}{\emph{AAAI}}. \bibinfo{pages}{689--696}.
\newblock


\bibitem[Yang and Gid{\'{o}}falvi(2018)]%
        {fmm}
\bibfield{author}{\bibinfo{person}{Can Yang} {and} \bibinfo{person}{Gy{\"{o}}z{\"{o}} Gid{\'{o}}falvi}.} \bibinfo{year}{2018}\natexlab{}.
\newblock \showarticletitle{Fast map matching, an algorithm integrating hidden Markov model with precomputation}.
\newblock \bibinfo{journal}{\emph{International Journal of Geographical Information Science}} \bibinfo{volume}{32}, \bibinfo{number}{3} (\bibinfo{year}{2018}), \bibinfo{pages}{547--570}.
\newblock


\bibitem[Yang et~al\mbox{.}(2022b)]%
        {GETNext}
\bibfield{author}{\bibinfo{person}{Song Yang}, \bibinfo{person}{Jiamou Liu}, {and} \bibinfo{person}{Kaiqi Zhao}.} \bibinfo{year}{2022}\natexlab{b}.
\newblock \showarticletitle{GETNext: Trajectory Flow Map Enhanced Transformer for Next {POI} Recommendation}. In \bibinfo{booktitle}{\emph{SIGIR}}. \bibinfo{pages}{1144--1153}.
\newblock


\bibitem[Yang et~al\mbox{.}(2021b)]%
        {pim}
\bibfield{author}{\bibinfo{person}{Sean~Bin Yang}, \bibinfo{person}{Chenjuan Guo}, \bibinfo{person}{Jilin Hu}, \bibinfo{person}{Jian Tang}, {and} \bibinfo{person}{Bin Yang}.} \bibinfo{year}{2021}\natexlab{b}.
\newblock \showarticletitle{Unsupervised Path Representation Learning with Curriculum Negative Sampling}. In \bibinfo{booktitle}{\emph{IJCAI}}. \bibinfo{pages}{3286--3292}.
\newblock


\bibitem[Yang et~al\mbox{.}(2022a)]%
        {yang2022wsccl}
\bibfield{author}{\bibinfo{person}{Sean~Bin Yang}, \bibinfo{person}{Chenjuan Guo}, \bibinfo{person}{Jilin Hu}, \bibinfo{person}{Bin Yang}, \bibinfo{person}{Jian Tang}, {and} \bibinfo{person}{Christian~S. Jensen}.} \bibinfo{year}{2022}\natexlab{a}.
\newblock \showarticletitle{Weakly-supervised Temporal Path Representation Learning with Contrastive Curriculum Learning}. In \bibinfo{booktitle}{\emph{ICDE}}. \bibinfo{pages}{2873--2885}.
\newblock


\bibitem[Yang et~al\mbox{.}(2023)]%
        {lightpath}
\bibfield{author}{\bibinfo{person}{Sean~Bin Yang}, \bibinfo{person}{Jilin Hu}, \bibinfo{person}{Chenjuan Guo}, \bibinfo{person}{Bin Yang}, {and} \bibinfo{person}{Christian~S. Jensen}.} \bibinfo{year}{2023}\natexlab{}.
\newblock \showarticletitle{LightPath: Lightweight and Scalable Path Representation Learning}. In \bibinfo{booktitle}{\emph{KDD}}. \bibinfo{pages}{2999--3010}.
\newblock


\bibitem[Yao et~al\mbox{.}(2022)]%
        {trajgat}
\bibfield{author}{\bibinfo{person}{Di Yao}, \bibinfo{person}{Haonan Hu}, \bibinfo{person}{Lun Du}, \bibinfo{person}{Gao Cong}, \bibinfo{person}{Shi Han}, {and} \bibinfo{person}{Jingping Bi}.} \bibinfo{year}{2022}\natexlab{}.
\newblock \showarticletitle{TrajGAT: {A} Graph-based Long-term Dependency Modeling Approach for Trajectory Similarity Computation}. In \bibinfo{booktitle}{\emph{KDD}}. \bibinfo{pages}{2275--2285}.
\newblock


\bibitem[Yao et~al\mbox{.}(2017)]%
        {traj2vec}
\bibfield{author}{\bibinfo{person}{Di Yao}, \bibinfo{person}{Chao Zhang}, \bibinfo{person}{Zhihua Zhu}, \bibinfo{person}{Jian{-}Hui Huang}, {and} \bibinfo{person}{Jingping Bi}.} \bibinfo{year}{2017}\natexlab{}.
\newblock \showarticletitle{Trajectory clustering via deep representation learning}. In \bibinfo{booktitle}{\emph{IJCNN}}. \bibinfo{pages}{3880--3887}.
\newblock


\bibitem[Zhang et~al\mbox{.}(2020)]%
        {traj2simvec}
\bibfield{author}{\bibinfo{person}{Hanyuan Zhang}, \bibinfo{person}{Xinyu Zhang}, \bibinfo{person}{Qize Jiang}, \bibinfo{person}{Baihua Zheng}, \bibinfo{person}{Zhenbang Sun}, \bibinfo{person}{Weiwei Sun}, {and} \bibinfo{person}{Changhu Wang}.} \bibinfo{year}{2020}\natexlab{}.
\newblock \showarticletitle{Trajectory Similarity Learning with Auxiliary Supervision and Optimal Matching}. In \bibinfo{booktitle}{\emph{IJCAI}}. \bibinfo{pages}{3209--3215}.
\newblock


\bibitem[Zhang et~al\mbox{.}(2023)]%
        {oasd}
\bibfield{author}{\bibinfo{person}{Qianru Zhang}, \bibinfo{person}{Zheng Wang}, \bibinfo{person}{Cheng Long}, \bibinfo{person}{Chao Huang}, \bibinfo{person}{Siu{-}Ming Yiu}, \bibinfo{person}{Yiding Liu}, \bibinfo{person}{Gao Cong}, {and} \bibinfo{person}{Jieming Shi}.} \bibinfo{year}{2023}\natexlab{}.
\newblock \showarticletitle{Online Anomalous Subtrajectory Detection on Road Networks with Deep Reinforcement Learning}. In \bibinfo{booktitle}{\emph{ICDE}}. \bibinfo{pages}{246--258}.
\newblock


\bibitem[Zhao et~al\mbox{.}(2022)]%
        {STGCN}
\bibfield{author}{\bibinfo{person}{Pengpeng Zhao}, \bibinfo{person}{Anjing Luo}, \bibinfo{person}{Yanchi Liu}, \bibinfo{person}{Jiajie Xu}, \bibinfo{person}{Zhixu Li}, \bibinfo{person}{Fuzhen Zhuang}, \bibinfo{person}{Victor~S. Sheng}, {and} \bibinfo{person}{Xiaofang Zhou}.} \bibinfo{year}{2022}\natexlab{}.
\newblock \showarticletitle{Where to Go Next: {A} Spatio-Temporal Gated Network for Next {POI} Recommendation}.
\newblock \bibinfo{journal}{\emph{IEEE Transactions on Knowledge and Data Engineering}} \bibinfo{volume}{34}, \bibinfo{number}{5} (\bibinfo{year}{2022}), \bibinfo{pages}{2512--2524}.
\newblock


\bibitem[Zhao et~al\mbox{.}(2023)]%
        {zhao2023traffic_prediction}
\bibfield{author}{\bibinfo{person}{Yusheng Zhao}, \bibinfo{person}{Xiao Luo}, \bibinfo{person}{Wei Ju}, \bibinfo{person}{Chong Chen}, \bibinfo{person}{Xian{-}Sheng Hua}, {and} \bibinfo{person}{Ming Zhang}.} \bibinfo{year}{2023}\natexlab{}.
\newblock \showarticletitle{Dynamic Hypergraph Structure Learning for Traffic Flow Forecasting}. In \bibinfo{booktitle}{\emph{ICDE}}. \bibinfo{pages}{2303--2316}.
\newblock


\bibitem[Zheng et~al\mbox{.}(2010)]%
        {geolife-dataset}
\bibfield{author}{\bibinfo{person}{Yu Zheng}, \bibinfo{person}{Xing Xie}, {and} \bibinfo{person}{Wei{-}Ying Ma}.} \bibinfo{year}{2010}\natexlab{}.
\newblock \showarticletitle{GeoLife: {A} Collaborative Social Networking Service among User, Location and Trajectory}.
\newblock \bibinfo{journal}{\emph{IEEE Data Engineering Bulletin}} \bibinfo{volume}{33}, \bibinfo{number}{2} (\bibinfo{year}{2010}), \bibinfo{pages}{32--39}.
\newblock


\bibitem[Zhou et~al\mbox{.}(2023)]%
        {grlstm}
\bibfield{author}{\bibinfo{person}{Silin Zhou}, \bibinfo{person}{Jing Li}, \bibinfo{person}{Hao Wang}, \bibinfo{person}{Shuo Shang}, {and} \bibinfo{person}{Peng Han}.} \bibinfo{year}{2023}\natexlab{}.
\newblock \showarticletitle{{GRLSTM:} Trajectory Similarity Computation with Graph-Based Residual {LSTM}}. In \bibinfo{booktitle}{\emph{AAAI}}. \bibinfo{pages}{4972--4980}.
\newblock


\bibitem[Zhu et~al\mbox{.}(2017)]%
        {time-lstm}
\bibfield{author}{\bibinfo{person}{Yu Zhu}, \bibinfo{person}{Hao Li}, \bibinfo{person}{Yikang Liao}, \bibinfo{person}{Beidou Wang}, \bibinfo{person}{Ziyu Guan}, \bibinfo{person}{Haifeng Liu}, {and} \bibinfo{person}{Deng Cai}.} \bibinfo{year}{2017}\natexlab{}.
\newblock \showarticletitle{What to Do Next: Modeling User Behaviors by Time-LSTM}. In \bibinfo{booktitle}{\emph{IJCAI}}. \bibinfo{pages}{3602--3608}.
\newblock


\end{thebibliography}

\end{document}